\documentclass[lettersize,journal]{IEEEtran}
\usepackage{amsmath,amsfonts}
\usepackage{algorithmic}
\usepackage{algorithm}
\usepackage{array}
\usepackage[caption=false,font=normalsize,labelfont=sf,textfont=sf]{subfig}
\usepackage{textcomp}
\usepackage{stfloats}
\usepackage{url}
\usepackage{verbatim}
\usepackage{graphicx}
\usepackage{cite}
\usepackage[normalem]{ulem}

\usepackage{lipsum} 
\usepackage{tabularx,colortbl}
\usepackage{float}
\usepackage[figurename=Fig.]{caption}
\usepackage{makecell}
\floatstyle{plaintop}
\restylefloat{table}

\usepackage{soul} 
\newcolumntype{?}{!{\vrule width 1pt}}
\newcolumntype{P}[1]{>{\centering\arraybackslash}p{#1}}
\newcolumntype{C}[1]{>{\centering\arraybackslash}c{#1}}
\usepackage[pagebackref=true,breaklinks=true,letterpaper=true,colorlinks,bookmarks=false]{hyperref}
\usepackage{url}

\usepackage{breakurl}
\usepackage{layouts}

\begin{document}

\title{Helping Visually Impaired People Take Better Quality Pictures}

\author{Maniratnam Mandal, Deepti Ghadiyaram, Danna Gurari, and Alan C. Bovik, ~\IEEEmembership{Fellow,~IEEE,}
\thanks{M. Mandal and A.C. Bovik are with the Laboratory for Image and Video Engineering, The University of Texas at Austin, Austin, TX, 78712, USA (e-mail: mmandal@utexas.edu; bovik@ece.utexas.edu).}
\thanks{D. Ghadiyaram is with Meta AI Research (email: deeptigp@fb.com).}
\thanks{D. Gurari is with The University of Colorado Boulder (email: danna.gurari@colorado.edu).}
}



\maketitle
\begin{abstract}
Perception-based image analysis technologies can be used to help visually impaired people take better quality pictures by providing automated guidance, thereby empowering them to interact more confidently on social media. The photographs taken by visually impaired users often suffer from one or both of two kinds of quality issues: technical quality (distortions), and semantic quality, such as framing and aesthetic composition. Here we develop tools to help them minimize occurrences of common technical distortions, such as blur, poor exposure, and noise. We do not address the complementary problems of semantic quality, leaving that aspect for future work.
The problem of assessing, and providing actionable feedback on the technical quality of pictures captured by visually impaired users is hard enough, owing to the severe, commingled distortions that often occur.
To advance progress on the problem of analyzing and measuring the technical quality of visually impaired user-generated content (VI-UGC), we built a very large and unique subjective image quality and distortion dataset. 
This new perceptual resource, which we call the LIVE-Meta VI-UGC Database, contains $40$K real-world distorted VI-UGC images and $40$K patches, on which we recorded $2.7$M human perceptual quality judgments and $2.7$M distortion labels. Using this psychometric resource we also created an automatic blind picture quality and distortion predictor that learns local-to-global spatial quality relationships, achieving state-of-the-art prediction performance on VI-UGC pictures, significantly outperforming existing picture quality models on this unique class of distorted picture data. We also created a  prototype feedback system that helps to guide users to mitigate quality issues and take better quality pictures, by creating a multi-task learning framework. 

\end{abstract}


\section{Introduction}
Computer vision solutions can lead to scaleable approaches for making technologies more accessible which, in turn, can contribute to community building. One example is making social media more accessible to visually impaired people. Being able to automatically understand picture and video content by AI-driven assistance could benefit low-vision/blind users when selecting pictures to upload on social media. While there has been progress on building tools to assist visually impaired users on other photography tasks \cite{vizwiz, OGvizwiz, scansearch, taptapsee}, studies \cite{visstudy1, visstudy2, visstudy3, visstudy4} have shown that such users still often rely on friends when taking pictures, leaving them feeling vulnerable and disempowered. These studies also have reported how visually impaired individuals often ask for information about and assistance with capturing high quality pictures. One way of helping these individuals to be more independent is to develop automatic picture quality raters that can supply guidance and feedback on the quality of pictures photographers are capturing, while also suggesting ways to ameliorate quality problems. 

\begin{figure}[t]
\begin{center}
\includegraphics[width=1\linewidth]{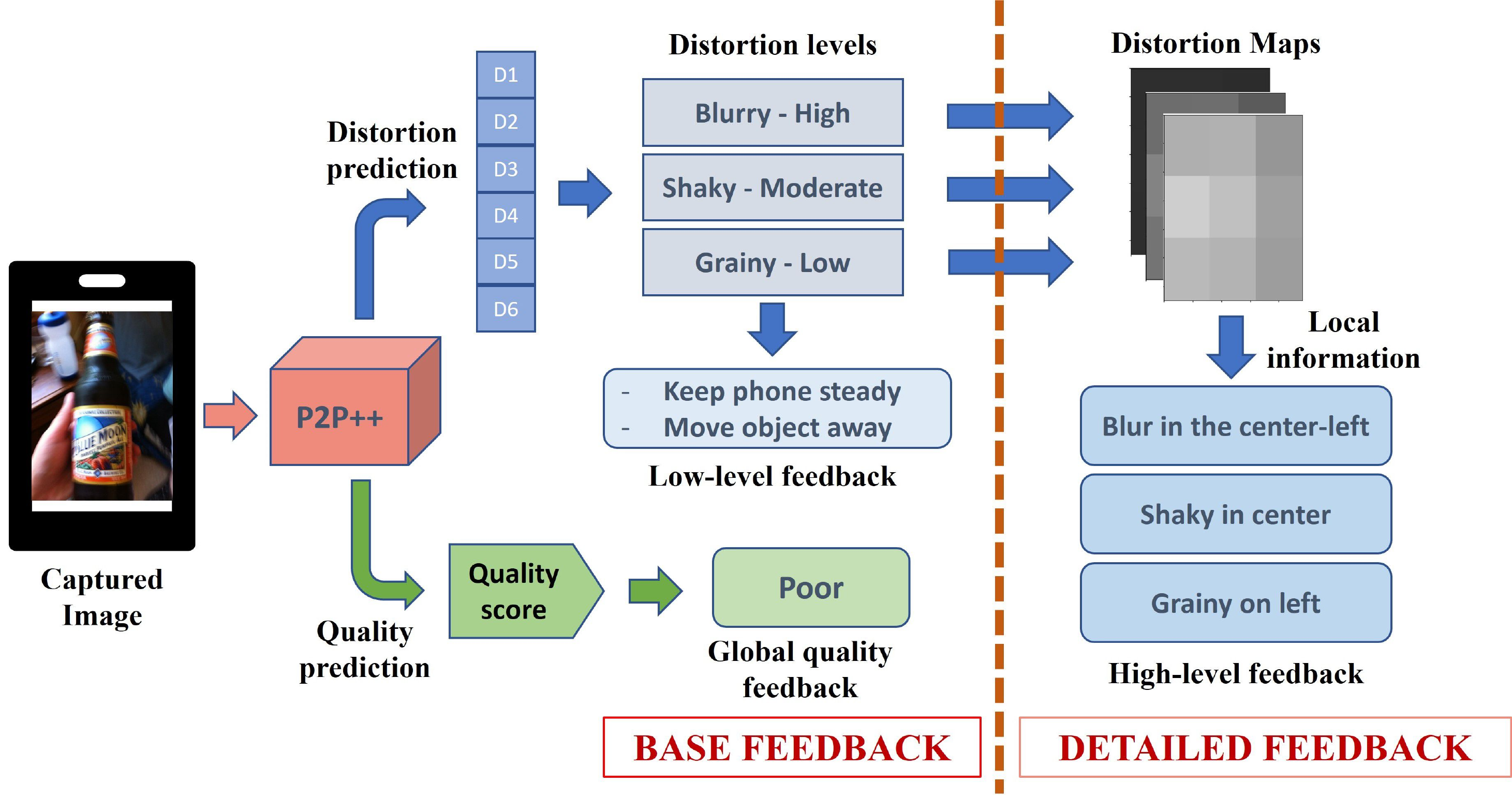}
\caption{\scriptsize{\textbf{Quality feedback to assist visually challenged users:} The captured image is passed through P2P++ (Sec.~\ref{sec:modeling}), which generates global quality and distortion scores. The predicted scores are used to provide suitable feedback on distortions present and ways to mitigate them. (Sec.~\ref{sec:modelExt})}}
\vspace{-2em}
\label{fig:teaser}
\end{center}
\end{figure}

A large number of models are available that can predict both perceptual quality (quality of content perceived by a user) and distortion types (blur, noise, underexposed, etc).  These models typically are based on finding perturbations of perceptually relevant `quality-aware' picture statistics \cite{brisque}. Yet, obtaining high prediction accuracy on \emph{user-generated content (UGC)} like that encountered on social media platforms remains a challenging problem \cite{paq2piq, nima, cnnIqa} due to the wide variation of commingled distortions which is hard to model.  Analyzing UGC pictures is difficult due to the absence of (pristine) reference pictures, leaving only No-Reference (NR) picture quality prediction models as suitable in such situations. However, existing state-of-the-art NR picture quality do not transfer well when applied to visually impaired user-generated pictures. We suspect part of the reason for the poor performance is because predictors were trained on datasets containing pictures captured by users without vision impairments, which is a domain shift from the quality issues observed in images taken by individuals with vision impairments.

Moreover, pictures that are captured by visually impaired users, without any guiding feedback, usually suffer from higher levels of distortion (blur, over/underexposure, etc.) \cite{vizwiz} as compared to users without vision impairments.


In this work, we aim to introduce a computer vision solution that can assist visually impaired people to take better pictures by providing feedback (e.g., auditory, haptic) describing distortions during picture capture and actions that may be taken to prevent them.\footnote{While pictures that are captured by visually impaired users may also suffer from semantic/aesthetic flaws, incorrect framing, or improper orientation \cite{assistframework}, \cite{easysnap}, and \cite{OGvizwiz}, we do not include this aspect of picture quality in the current study, owing to the considerable difficulty in conducting massive-scale human studies on both aspects at the same time. However, we do recognize that photographic semantics/aesthetics are also important to address for a comprehensive picture quality monitoring and feedback system.} Of note, distortions may arise in many different forms because of imperfect capture devices, focusing issues, stabilization problems, camera movement, sub-optimal lighting, and other sources.  Such distortions can potentially be amplified for visually impaired photographers, since they cannot inspect their pictures to verify quality. Moreover, multiple distortions often intermix, making them harder to separately identify and classify. 


Towards creating accurate predictors of technical picture quality, we develop a large psychometric dataset of VI-UGC pictures labeled by many human participants.  We call this new dataset the LIVE-Meta VI-UGC Database.  This new dataset supports creating, benchmarking, and comparing picture quality models. Moreover, it fills a gap with prior work.  For example, while a variety of public image quality datasets are available that have driven the development of NR-IQA (no-reference image quality assessment) predictors for naturalistic pictures \cite{paq2piq, nima}, most do not contain pictures taken by visually impaired individuals.  Yet, pictures from visually impaired individuals are different in important ways. One dataset does contain labels of picture distortion categories for VI-UGC: VizWiz-QualityIssues \cite{vizwiz}.  However, it only provides categories labels for the global images rather than also for localized regions in the images where such quality issues can occur.  Like prior work, our proposed dataset also extends the large VizWiz dataset, which consists of images taken by visually impaired photographers.  In contrast, we conducted a large-scale visual psychometric study on both the images and extracted patches (randomly selected and salient patches), whereby we collected human subjective quality scores and distortion labels. Our work is inspired in part by the recent finding on perceptual QA \cite{paq2piq, patchvq} that modeling the relationship between local and global distortions can lead to better visual quality predictions.  We are sharing this dataset online for public use.   



We next developed models using our large scale dataset as a promising foundation for a feedback system that can assist visually impaired users to take higher quality pictures. Our proposed models illustrate that using local quality and distortion information can lead to accurate global and local predictions with fewer parameters.

In summary, the contributions of this work are as follows: 

\begin{itemize}
\item \textbf{We built the largest subjective image quality and distortion database targeting pictures captured by visually impaired users.} This new resource contains about $40$K images collected from VizWiz \cite{vizwiz} and $40$K patches (half randomly selected and half salient) cropped from them. We conducted a large-scale subjective picture quality study on them, collecting $2.7$M labels of both perceived quality and judged distortion types. The new LIVE-Meta VI-UGC dataset is also the largest publicly available distortion classification dataset. Further, as a control, we also collected about $7$5K ratings on $2.2$K frames from videos captured by visually impaired photographers, which are provided in the ORBIT \cite{orbit} dataset (Sec.~\ref{sec:dataset_humanstudy}).
\item \textbf{We created a state-of-the-art blind (or \textit{no-reference}) VI-UGC picture quality and distortion predictor.} Using a deep neural architecture based on the recent successful PaQ-2-PiQ model \cite{paq2piq}, we created a multi-task learning system that is able to predict both the perceptual quality of pictures captured by visually impaired users, and the possible presence of five common picture distortions. Since the model is trained on patches, it is able to predict spatial maps of both quality and distortion types. We implemented the new prediction model as an algorithm that we will refer to as P2P++. It achieves top performance on the new dataset and on the independent dataset of ORBIT images, as compared to other NR-IQA models (Sec.~\ref{sec:modeling} and \ref{sec:cross_data}).
\item \textbf{Using the multi-task model, we also created a prototype feedback system to assist visually impaired users to take better quality pictures.} The P2P++ system provides feedback on overall (global) picture quality, along with suggestions on how to mitigate quality issues. This discussion includes ways by which the obtained spatial distortion maps can be used to generate detailed, localized feedback, and realized these ideas in actual smartphone (iOS and Android) implementations (Sec.~\ref{sec:feedback}).
\end{itemize}

\section{Related Work}\label{sec:background}
\subsection{Image Quality and Distortion Datasets:}
The most heavily-used datasets in image quality research remain older corpora of synthetically applied distortions of natural pictures. Since synthetic distortions are quite different from authentic, real-world distortions, NR-IQA models trained on them perform poorly on real-world distorted content \cite{clive}. Such ``legacy" datasets include LIVE \cite{hamidLiveDB}, CSIQ \cite{csiqdata}, TID-2008 \cite{ponomarenko2009tid2008}, and TID-2013 \cite{tid2013}. These contain small numbers ($< 30$) of unique picture contents that have been synthetically distorted by applying single quality impairments (JPEG, Gaussian blur, etc.). As such, they do not capture the extremely diverse, complex mixtures of distortions that arise in real-world settings. 
It is important to understand that the degree of technical distortion is not the same as perceived quality, since the latter is also deeply affected by the content and by perceptual processes such as masking \cite{mseLoveLeave, vpooling}. Accomplishing both tasks is both important and more difficult.

More recent datasets such as Live Challenge \cite{clive}, KonIQ \cite{koniq}, and LIVE-FB \cite{paq2piq} have tried to address these problems by obtaining subjective labels on much larger numbers of UGC images affected by real-world distortions. While these resources contain subjective quality labels, they do not include labels of distortion types, except for the smaller LIVE Challenge dataset \cite{clive}, which contains both quality and distortion labels on about $1200$ images captured with mobile devices.  The Flickr-Distortion dataset \cite{flickrdist} contains synthetic distortion labels on $804$ Flickr UGC images, but it does not contain any quality labels, and it has not been made public. The VizWiz-QualityIssues \cite{vizwiz} dataset is the only available resource that contains images taken by visually impaired users. It also contains distortion labels for some common types of quality impairments and aesthetic flaws. However, since it lacks picture quality scores and supplies only 5 human distortion classification ratings per image, it cannot be used for our purposes. 
The lack of any substantial UGC dataset (captured by visually impaired users or otherwise) containing large numbers of both quality scores and distortion labels motivated us to conduct a large-scale psychometric study of the perceptual quality and distortions present in typical VI-UGC pictures. The dataset we created is $33$ times larger than any prior picture quality and distortion labeled dataset, and it is the only dataset of this kind that targets the area of assisting visually impaired photographers. 

\subsection{Image Quality Models}
For this application, full-reference (FR) image quality models like SSIM \cite{ssim} are of no use, since they require the availability of pristine reference images to predict relative picture fidelity. Instead, we are constrained to consider NR (blind) models, several of which have achieved success. Popular NR picture quality prediction models include BRISQUE \cite{brisque}, NIQE \cite{niqe}, CORNIA \cite{cornia}, and FRIQUEE \cite{friquee}, which use ``handcrafted" statistical features to train shallow learners (typically support vector regressors, or SVRs) to predict picture quality. While these models work well on legacy single synthetic distortion datasets \cite{hamidLiveDB, tid2013, ponomarenko2009tid2008} they perform poorly on real-world, UGC data \cite{koniq, paq2piq}. Deep NR-IQA models \cite{ghadiyaram2014blind, deepConvIQA, bosseDeepIQA, fullyDeepIQA} have also been developed that also perform quite well on synthetic distortion datasets, but still struggle on UGC datasets like LIVE-Challenge \cite{clive} and LIVE-FB \cite{paq2piq}. An advanced deep model called PaQ-2-PiQ \cite{paq2piq} leverages relationships between local and global quality predictions to achieve SOTA performance on the largest and most comprehensive UGC picture quality datasets. A few multi-task models like IQA-CNN++ \cite{cnniqapp} and QualNet \cite{qualnet} are available that use relationships between quality and distortion features to predict both picture quality scores and distortion categories. These multi-task models also perform well on synthetic datasets, but struggle on real-world UGC pictures and distortions. The authors of \cite{vizwiz} used an Xception backbone pre-trained on ImageNet \cite{imageNet} to predict distortion categories on the VizWiz pictures, achieving promising results.   

\subsection{Assisting the Visually Impaired}
Making visual media accessible to visually impaired users has been a long standing problem. For example, earlier work attempted to convert visual signals to be perceived into tactile forms to mitigate the lack of sight \cite{barner1}, \cite{barner2}, \cite{barner3}. Over the last decade, several applications have been developed to help visually impaired users capture better images. Most of these applications were built for visual recognition tasks \cite{taptapsee}, \cite{easysnap}, \cite{OGvizwiz}. \textit{TapTapSee} \cite{taptapsee} helps a user take focus-adjusted images, whereas applications like \textit{VizWiz Ver2} \cite{OGvizwiz} and \textit{EasySnap} \cite{easysnap} use simple darkness and blur detection algorithms. All of these make simple distortion measurements to assist automated object recognition. The authors of \cite{assistframework} developed an assisted photography framework to help users better frame their photos, using an image composition model that assesses aesthetic quality. The \textit{Scan Search} \cite{scansearch} application uses the Lucas-Kanade \cite{lucaskanade} optical flow method to track feature points and determine the stability of the camera. The most stable frames are used for online object detection. Both provide automatic selection of a best image. The authors of \cite{vizwiz} also developed algorithms to detect the recognizability and answerability of images captured by blind users, for image captioning and visual question answering applications. None of these efforts were predicated on perceptual models, perceptual quality, or were trained on human data. The absence of any robust model that can predict distortions that commonly afflict VI-UGC pictures, and that can be used to monitor and guide perceptual quality during capture motivated us to create a system capable of doing so.

\begin{figure}[b]
\vspace{-1em}
\begin{center}
\includegraphics[ width=1\linewidth]{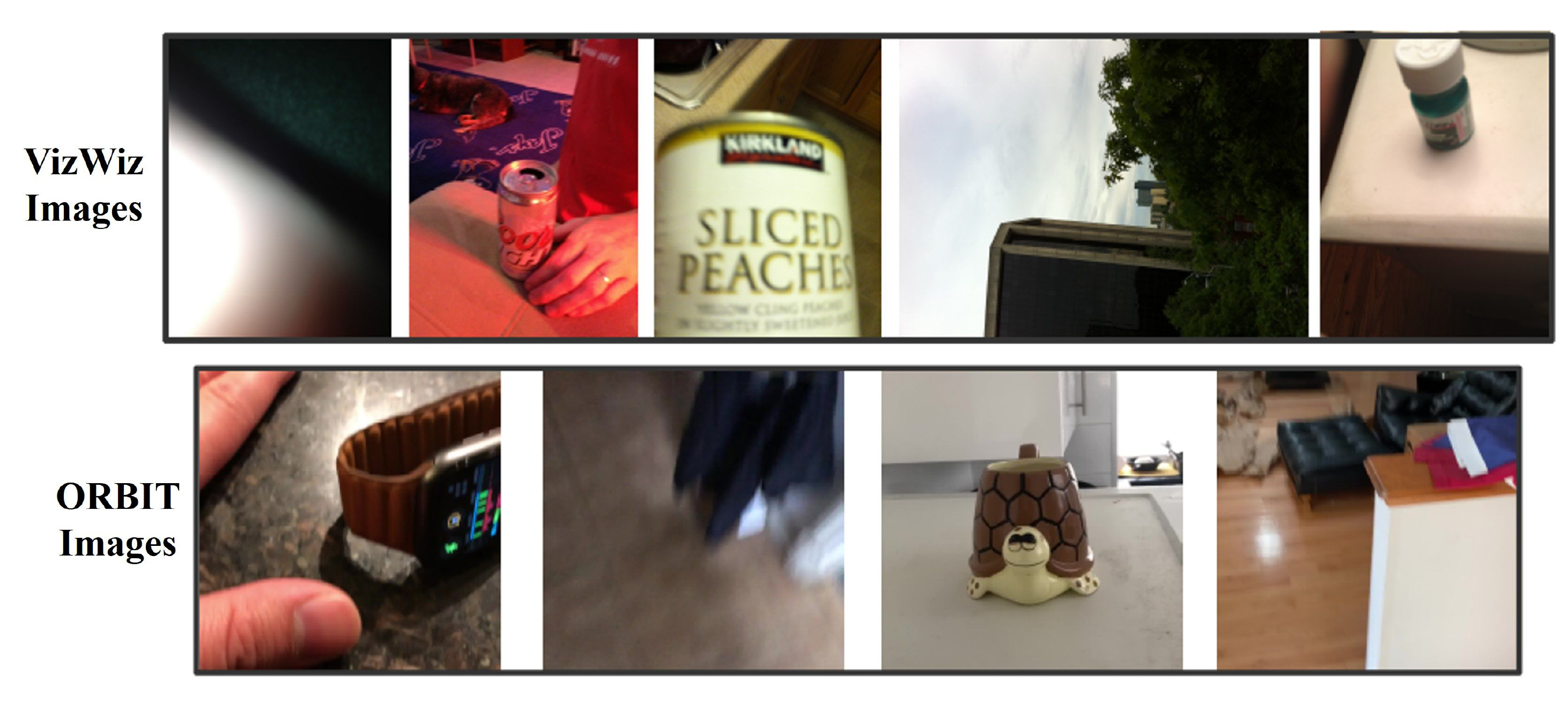}
\vspace{-1em}
\caption{\scriptsize{\textbf{Sample images from the two datasets - VizWiz (top row) and ORBIT (bottom row)}, each resized to fit. The actual images are of highly diverse sizes and resolutions.}}
\label{fig:exemplarFLIVE}
\end{center}
\end{figure}
\section{Dataset and Human Study}\label{sec:dataset_humanstudy}
This section elaborates the construction of the new LIVE-Meta VI-UGC dataset and details the online human study by which we collected subjective quality labels. The proposed dataset contains $39,660$ images, along with $39,660$ patches extracted from them, half of which are salient patches, and the other half randomly cropped. We also collected $2.7$M ratings and distortion labels on the images, and equal numbers on the patches. This dataset is much larger than any previous legacy synthetic dataset \cite{hamidLiveDB, tid2013, ponomarenko2009tid2008}, and significantly larger than any UGC dataset containing both quality and distortion labels \cite{clive}. It is also the first quality-focused dataset dedicated towards developing assistive technology for visually impaired photographers.    

\subsection{Constructing the Dataset}\label{sec:dataset}
\subsubsection{Building on VizWiz}
\begin{figure}[h]
\begin{center}
\includegraphics[width=0.8\linewidth]{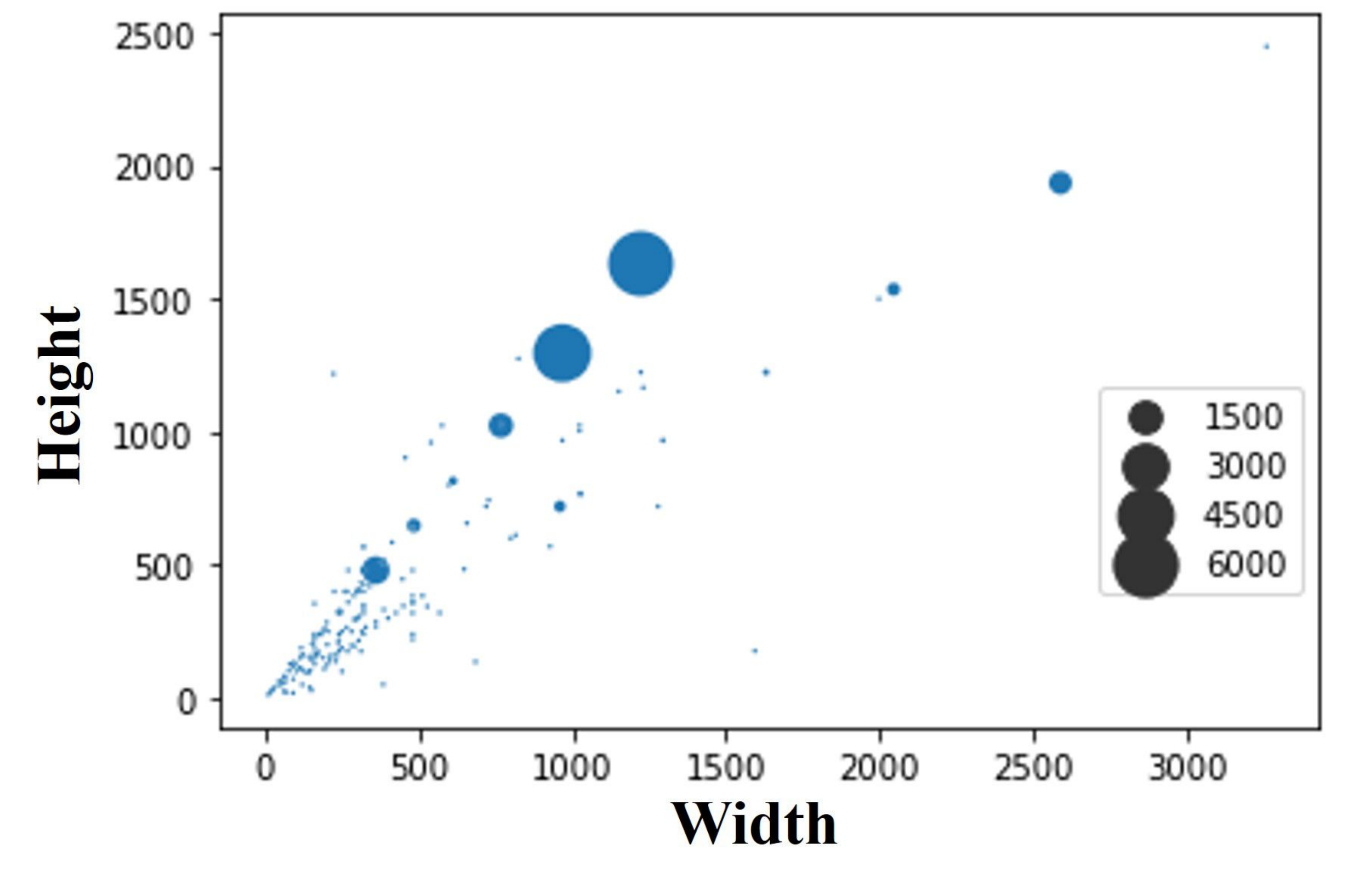} 
\caption{\scriptsize{\textbf{Left:} Scatter plot of picture sizes (width vs. height), where the circle sizes code the number of pictures of each size in the VizWiz database.}}
\label{fig:pixel_aspect}
\end{center}
\end{figure}

VizWiz \cite{Gurari2018VizWizGC} was the first publicly available dataset containing content generated by blind users. The images in the dataset (Fig. \ref{fig:exemplarFLIVE}) were generated under real use conditions via the VizWiz mobile application \cite{OGvizwiz}. Fig. \ref{fig:pixel_aspect} shows the diversity in resolution and aspect ratio of the dataset. Since the main purpose of the application was to submit images for image captioning and to answer visual questions, it addressed a practical need for visually impaired users, and is a suitable platform on which to develop quality-perceptive assistive technology. The authors of \cite{vizwiz} used the same dataset to collect some common quality (blur, exposure) and aesthetic labels (framing, obstruction, orientation) along with information on image recognizability. To the best of our knowledge, this was the first attempt to develop a dataset of this kind. A deeper analysis of the data collected revealed important insights on VI-UGC images. The authors considered blur and framing to be the most prevalent flaws of VI-UGC content that often render images unrecognizable. They also carried out a detailed analysis on the correlations between the different impairments, the relationships between recognizability and distortions, and pointed out crucial directions that research in this area should progress.

However, that work did not address measurement and analysis of the perceived qualities of VI-UGC pictures, which we have addressed by developing a large new dataset, on which we created VI-UGC specific IQA algorithms. These models address technical distortions such as blur, over/under exposure, and noise, but not semantic/aesthetic flaws such as framing, mood, and content selection. While the latter are important, they involve different capture problems and perceptual processes and should be treated differently \cite{paq2piq}, including using different kinds of training data. 

\subsubsection{Categorizing Distortions}
As mentioned previously, natural distortions are extremely diverse and commingle with each other, making it hard to exhaustively categorize them \cite{clive, koniq}. Since the images in our dataset were captured by visually impaired persons, there is a prevalence of heavy distortions. Among these, we chose to focus on the most common and identifiable ones. This resulted in five major categories: focus blur (`blurry'), motion blur (`shaky'), overexposure (`bright'), underexposure (`dark'), and noise (`grainy'). We also included two other categories: `none' (signifying the absence of any distortion) and `other' (distortions that are non-identifiable or cannot be categorized as being one of the other options).

\subsubsection{Cropping Patches}
\begin{figure}[h]
\begin{center}
\includegraphics[width=1\linewidth]{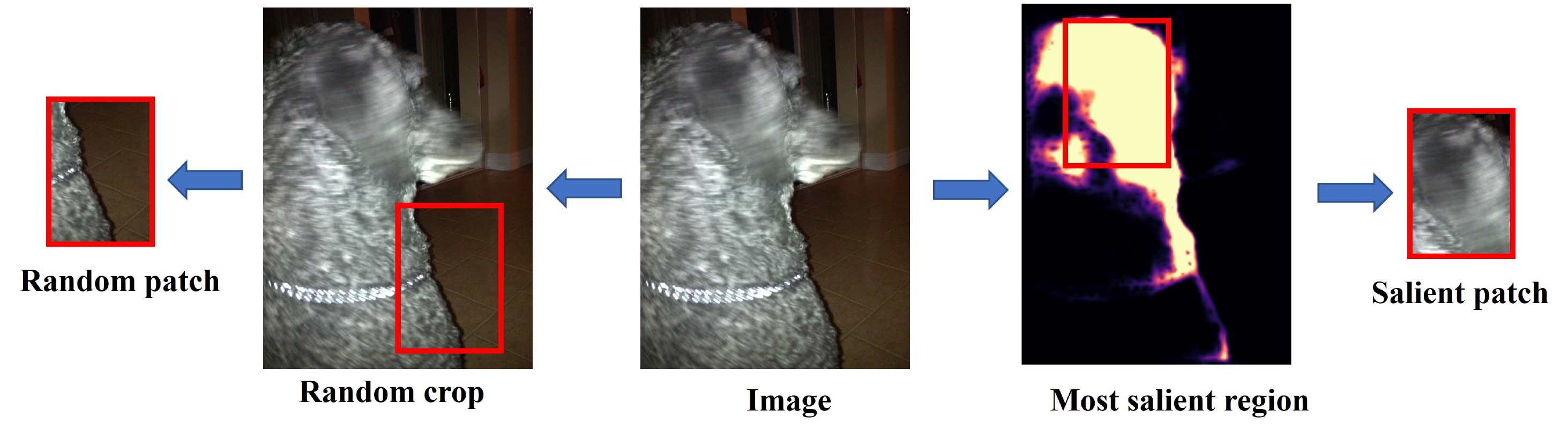}
\caption{\scriptsize{\textbf{Two kinds of spatial patches} were cropped from images, all to $40\%$ of the original image dimensions: randomly selected and salient patches cropped from disjoint halves of the overall image corpus. }}
\label{fig:egPatches}
\end{center}
\end{figure}

The relationships between local and global spatial quality have been shown to be important and, when modeled, to lead to improved quality predictions \cite{paq2piq, patchvq}. Here, we carry these ideas further by studying the impact of the method of choosing patches on quality prediction. To do this, we divided the entire dataset into two random halves. On half of the images, each randomly selected patch was cropped to $40\%$ of each of its linear spatial dimensions. On the other half of the images, a most visually salient patch of the same ($40\%$) dimensions was cropped. The SOTA pyramid feature attention network \cite{pyramidsal} was used to create saliency maps, from which the most salient patch was cropped. All of the patches have the same aspect ratios as the original image they were cropped from (Fig. \ref{fig:egPatches}).

\subsubsection{ORBIT Data}
To study the cross-dataset performance of our model and its applicability to video tasks, we also created a separate, smaller dataset containing images extracted from the ORBIT video database \cite{orbit}. The ORBIT dataset includes videos captured by visually impaired users, with the identification of common objects as the objective, under two environmental conditions -  clean and cluttered. We selected $59$ types of objects at random, and from each we selected two videos (one clean and one cluttered). We then sampled each video at one frame per second, collecting a total of $2,235$ frame images in total (Fig. \ref{fig:exemplarFLIVE}). Each image in the ORBIT sub-dataset has the same spatial dimensions as the original videos ($1080 \times 1080$). As before, we then collected global quality ratings on these to form the auxiliary LIVE-ORBIT frame quality dataset.

\subsection{Subjective Quality and Distortion Study}\label{sec:human_study}
Amazon Mechanical Turk (AMT) is a well-established crowdsourcing platform for subjective quality studies \cite{paq2piq, patchvq, vizwiz, clive, livevqc}. Our human study was carried out in three stages of data collection - image, patch, and ORBIT sessions. Overall, $3,945$ subjects participated in the study and, after rejection and cleaning, we collected an average of about $34$ ratings on each image and each patch. Our study was accessible to all platforms and geographical locations.

\subsubsection{AMT Study Design}

\begin{figure}[t]
\begin{center}
\includegraphics[width=\linewidth, trim={0em 0em 1em 0em},clip]{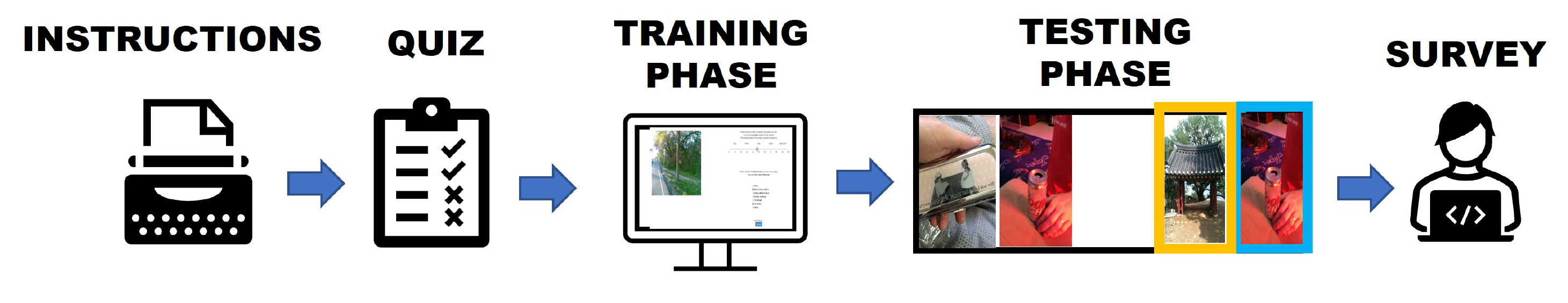}
\vspace{-2em}
\caption{\scriptsize{\textbf{Study workflow} for both image and patch sessions.}}
\vspace{-2em}
\label{fig:AMTdesign}
\end{center}
\end{figure}

The study workflow is shown in Fig. \ref{fig:AMTdesign}. The subjects were asked to participate in two tasks - image quality rating and distortion type identification. 
The AMT study started with a series of instructions, followed by a quiz, and then the training and testing phases. Before accepting the task, the workers could read a brief summary on the introductory page. Having accepted, they then had to read a second introductory page on distortions, then a set of timed instruction pages: Rating Instructions, Phase Instructions, Additional Instructions, and Ethics Policy. While they read the instructions, the subjects were screened based on the study criteria described in Sec. \ref{sec:study_requirements}. Following the instruction pages, they then had to pass a quiz in order to proceed to the training and testing phases. The training phase contained five sample distorted images to familiarize the subjects with the task. After training, the subjects entered the testing phase, where each subject rated $110$ images. Afterwards, each subject completed their task by answering a questionnaire regarding the study conditions and their demography.  

\begin{figure}[htb]
\begin{center}
\includegraphics[width=1\linewidth]{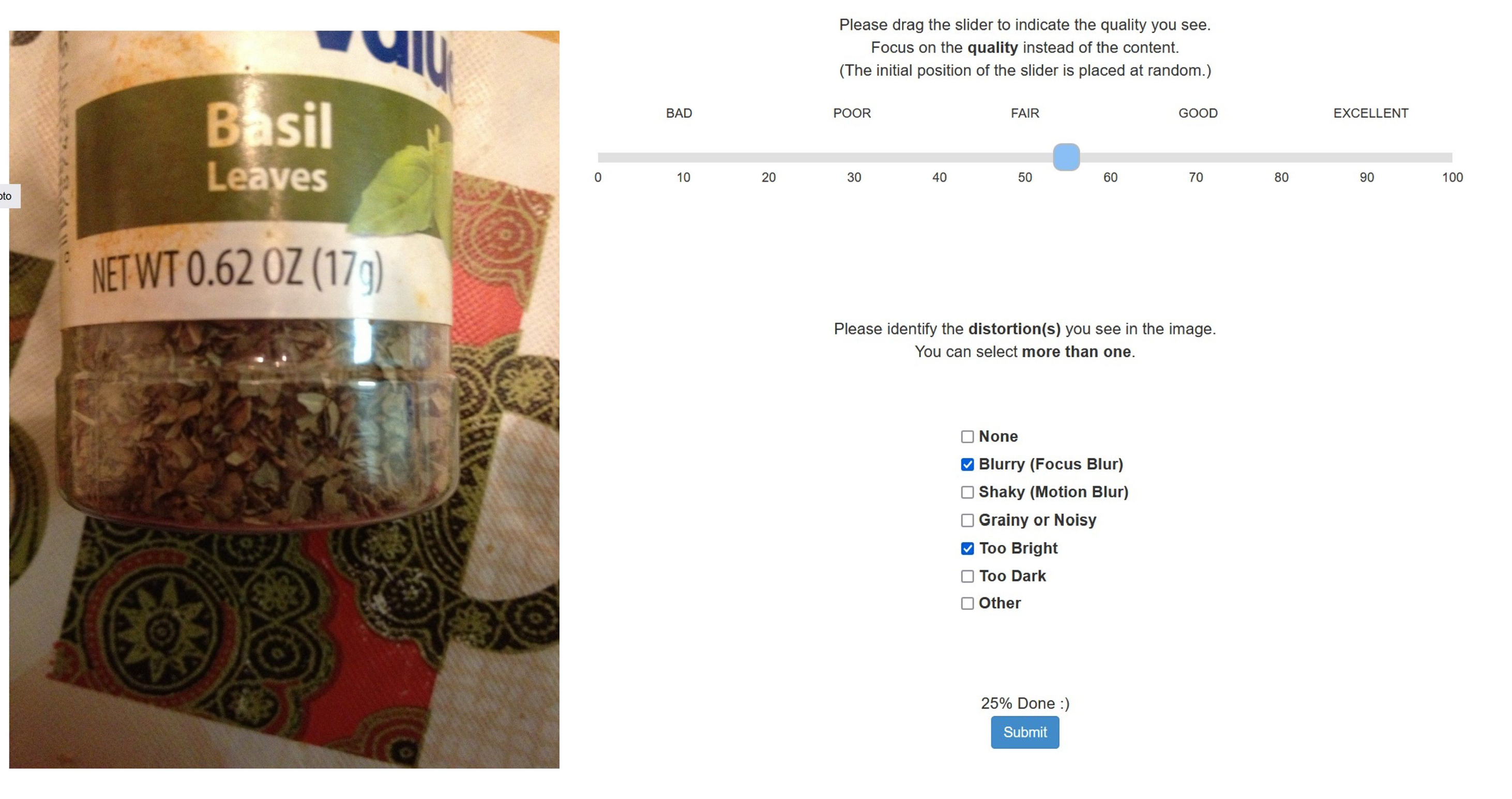}
\vspace{-1em}
\caption{\scriptsize{\textbf{Rating Interface} }}
\label{fig:interface_rating}
\end{center}
\end{figure}

The rating interface that the human subjects deployed is shown in Fig. \ref{fig:interface_rating}. The image was displayed on the left, and the rating section was on the right. The top half of the rating section displays the rating bar, which allowed the subjects to tender overall quality scores on a continuous scale from $0-100$ using a sliding cursor. The subjects were asked to provide their ratings anywhere they felt appropriate, guided by Likert scale \cite{likert} markings \textit{BAD}, \textit{POOR}, \textit{FAIR}, \textit{GOOD}, and \textit{EXCELLENT}. On the bottom half, they were asked to choose the distortion(s) that they deemed present in the image, from among the seven options discussed earlier (Sec \ref{sec:dataset}).

\subsubsection{Study Requirements} \label{sec:study_requirements}
Each subject viewed either only images, or only patches. For both image and patch sessions, each batch was published in four phases. The workers had to satisfy the following criteria to be eligible for the study: 
\begin{itemize}
\item \textbf{Worker Reliability: } Only accepted workers with an approval rate of $> 75\%$, with $>1000$ HITs approved.
\item \textbf{Browser Resolution: } $> 480$p for mobile devices, and $> 720$p for others.
\item \textbf{Browser Versions: } Latest versions of Chrome, Firefox, Edge, or Safari.
\item \textbf{Browser Zoom: } Set to $100\%$.
\end{itemize}

\subsubsection{Subject Rejection}
As has been discussed in previous studies~\cite{paq2piq, patchvq, vizwiz, clive, livevqc, koniq, ytugc}, online crowdsourcing carries the risk of distracted, inadequately equipped, disengaged, or even frankly dishonest subjects, so there is often a high percentage of unreliable labels. We used various strategies to screen the subjects using criteria applied both during and after the study. 

\textbf{During Task:} During the instruction phase, we checked whether the subject's browser window, browser and OS version, and zoom (non-magnified) condition satisfied the requirements stated in the instructions. If they did not, their participation was ended. To detect dishonest workers, at the halfway point of each testing phase, we processed the scores already given to determine whether they had been giving unchanging ratings (only nudging the slider / supplying haphazard scores) on either task, and were rejected accordingly. 

\textbf{Post Task:} Of the $110$ images viewed in a session, $5$ were randomly repeated. A subject was rejected if their ``repeat" scores were not consistently similar to the scores given the first time. We also included $5$ images from the LIVE-FB dataset \cite{paq2piq} designated as the ``golden" set, and removed the subjects if their ratings did not adequately match the golden ones. Overall, we rejected the scores given by $814$ subjects. 

\subsubsection{Data Cleaning}
The remaining scores after subject rejection were processed by a series of data cleaning steps: \textbf{(1)} removed $43$ images ($1.3$K ratings) of a constant value. \textbf{(2)} removed the ratings provided by subjects who did not wear their prescribed lenses ($0.9\%$ of total ratings removed).
\textbf{(3)} applied the ITU-R BT.500-14 \cite{itu_standard} (Annex 1, Sec 2.3) subject rejection protocol to screen $56$ more outlier subjects. \textbf{(4)} For each image and patch, we also rejected outliers from the individual score distributions, as follows. We first calculated the kurtosis \cite{kurtosis} to determine the normality of the scores. If they were determined to be normal, the Z-score outlier rejection method \cite{modzscore} was applied. Otherwise, the Tukey IQR method \cite{tukey1977exploratory} was applied. Overall, including all subject and score outlier rejections, around $1.7\%$ of the ratings were tossed out. We were left with about $2.7$M subject scores ($1.36$M on images, $1.33$M on patches) on VizWiz images, and $76$K ratings on the ORBIT images.

\section{Subjective Data Processing and Results}\label{sec:data_analysis}
\subsection{Subject Statistics}\label{sec:subjec_demography}
In the exit survey, the subjects were asked to provide information regarding the following: the display, viewing distance, gender, age, and whether the worker needed corrective lenses, and if so, whether he/she wore them during the study.
\subsubsection{Demographic Information} 
\begin{figure}[htb]
\begin{center}
\includegraphics[width=1\linewidth]{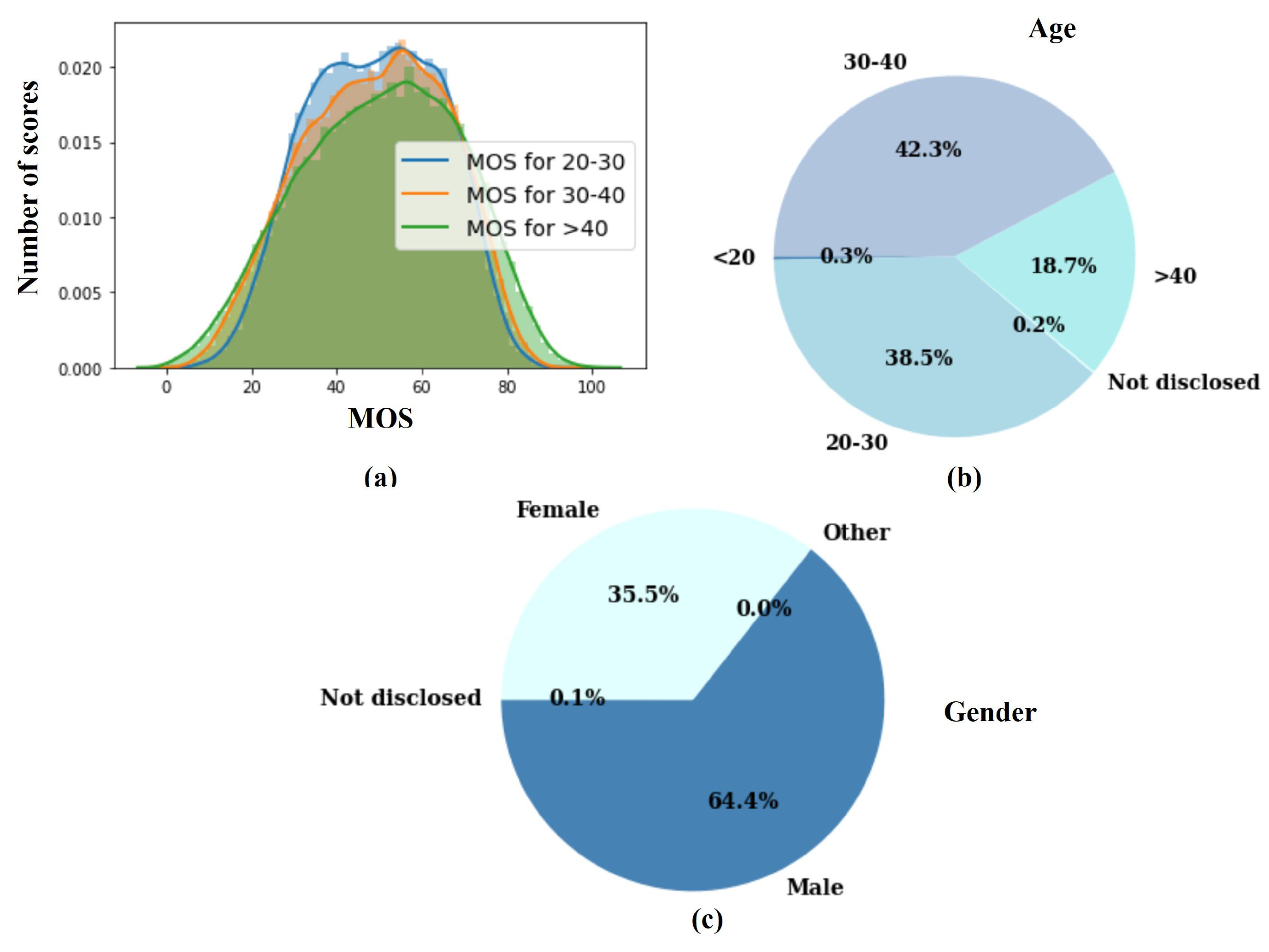}

\caption{\scriptsize{\textbf{Demographic information statistics} (a) MOS distribution for different age groups, (b) Age distribution, and (c) Gender distribution of the subjects who participated in the study.}}
\label{fig:demography}
\end{center}
\end{figure}

We did not put any regional qualification on subject participation. Fig. \ref{fig:demography}(a) shows the gender distribution with 64.4\% male and 35.5\% female participants. Fig. \ref{fig:demography}(b) shows the age distribution with 20-40 being the primary age group. From the age-specific MOS distribution (Fig. \ref{fig:demography}(c)), we observe that the standard deviation of the score distribution roughly decreases with age. This might be due to the younger demographic being exposed more to online user-generated digital media which has a narrow distribution, making them more tolerant to changes in quality.

\subsubsection{Viewing Conditions} 
\begin{figure}[htb]
\begin{center}
\includegraphics[width=1\linewidth]{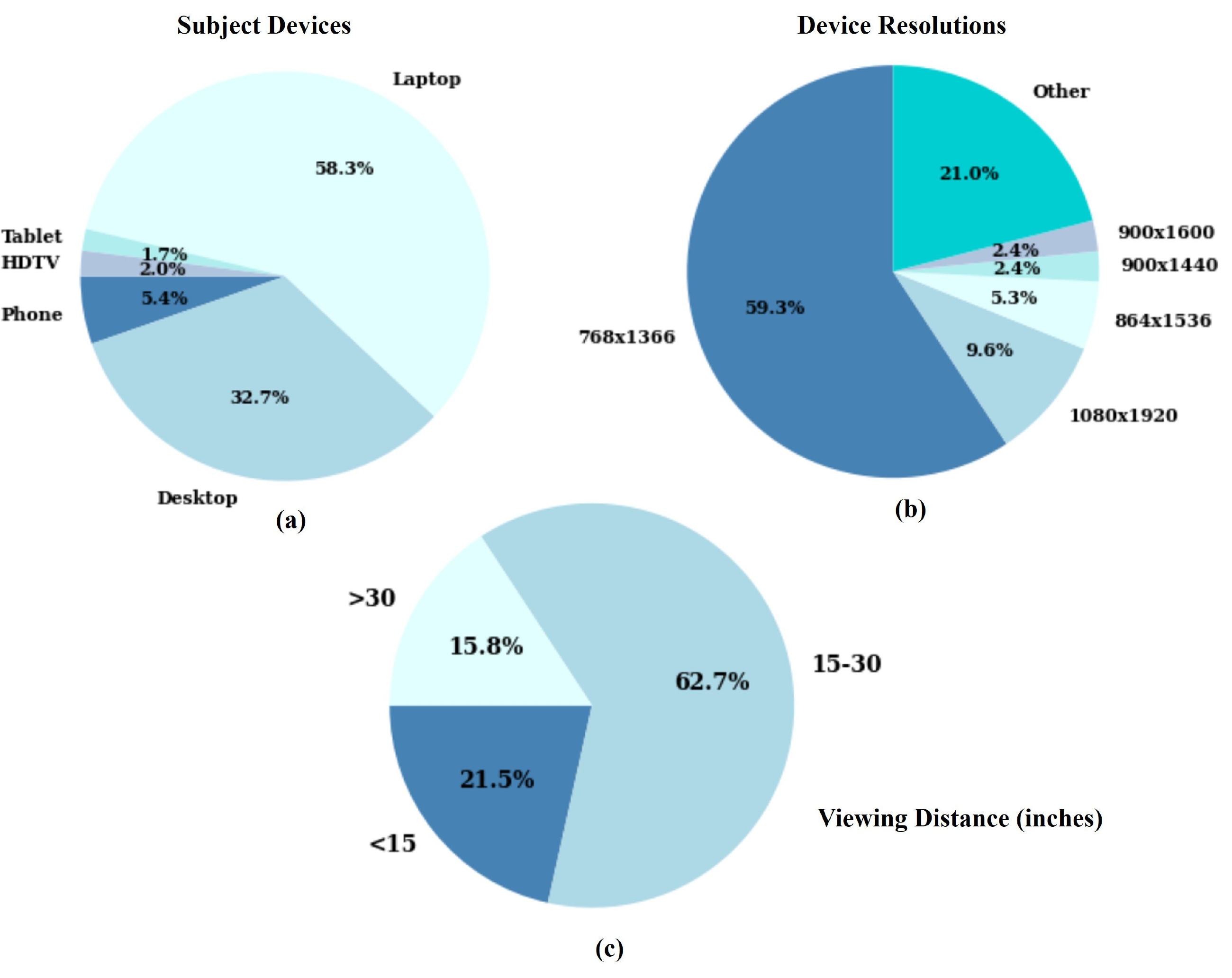}

\caption{\scriptsize{\textbf{Viewing condition statistics} (a) Devices and (b) Device Resolutions used by the subjects participating in the study.}}
\label{fig:deviceInfo}
\vspace{-1.5em}
\end{center}
\end{figure}

Due to the nature of crowdsourcing, subjects participated under diverse viewing conditions, including at different geographies, viewing distances, display devices, browsers, resolutions, ambient lighting, etc. Fig. \ref{fig:deviceInfo} shows the statistics of some of the viewing conditions. Laptops (58.3\%) and desktop workstations (32.7\%) were the most used devices (Fig. \ref{fig:deviceInfo}(a)), and most subjects (62.7\%) preferred 15-30 inches as the viewing distance (Fig. \ref{fig:deviceInfo}(c)). The most common screen resolution was $768\times1366$ (Fig. \ref{fig:deviceInfo}(b)), and $640\times360$ was the most common among mobile devices. 

\subsection{Data Analysis}
\subsubsection{Inter-subject quality consistency} An inter-subject consistency test~\cite{paq2piq, patchvq} was carried out by randomly dividing the subject pool into two disjoint sets of equal size, then calculating the Spearman Rank Correlation Coefficient (SRCC) \cite{kendall1948rank} between the two corresponding sets of MOS (mean opinion score). The average SRCC over $50$ such random splits yields a useful measure of inter-subject consistency. The average SRCC on VizWiz images was 0.93, on patches was 0.90 (0.87 on random and 0.92 on salient patches), and on ORBIT was 0.93. 
Among the patches, the average SRCC on the random patch set was 0.87, but for the salient patches was 0.92, hence the ratings on salient patches were more consistent. 
These results substantially validate the efficacy of our data collection and subject rejection processes.

\subsubsection{Intra-subject quality consistency} Each subject rated 5 ``golden" images. 
We computed the Linear Correlation Coefficient (LCC) \cite{pcc} between the mean of the ratings on the 5 ``golden" images with the original scores. The median PCC value over all subjects was 0.90 for the VizWiz image study, 0.90 for the patch study, and 0.87 for the ORBIT study. Again, these high correlations serve to validate our overall subjective study protocol.

\subsubsection{Patch vs image quality} 
\begin{figure}[htb]
\begin{center}
\includegraphics[width=\linewidth]{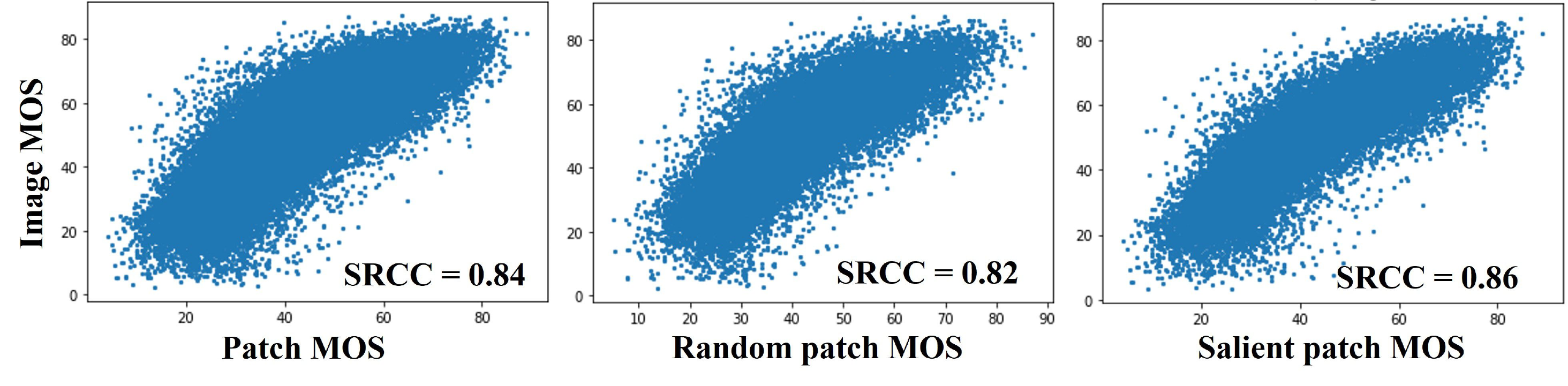}
\caption{\scriptsize{\textbf{Scatter plots of patch vs image MOS correlations.} Image MOS vs all patches (left), random patch (middle) and salient patch (right) MOS cropped from the same image.}}
\label{fig:patchCorrel}
\vspace{-1em}
\end{center}
\end{figure}

Fig. \ref{fig:patchCorrel} shows scatter plots of image MOS against patch MOS, for both kinds of patches. The SRCC obtained between image and patch MOS was 0.84, indicating a strong relationship between local and global image quality. The SRCC between image MOS and random and salient patch MOS was 0.82 and 0.86, respectively, suggesting that salient patches may play a strong role in the perception of global picture quality. This may be because some distortions are salient, and/or that distortions on salient regions are more annoying. 

\subsubsection{Quality rating consistency among subject demographics} We studied the effects of different parameters, associated with the study environment, on MOS. The SRCC of quality ratings collected on laptops and desktops (collectively accounting for $91\%$ of the devices used) against MOS was 0.91, while the SRCC of ratings provided on phones and other devices was 0.7. These results strongly suggest that perceptual quality is affected by the size of the display. We also studied other parameters by computing correlations between two major resolutions: $768\times 1366$ and $1080\times 1920$ (0.87); between two major viewing distances; $<$ 15 and 15-30 inches (0.89); major age groups: 20-30 and 30-40 (0.91); and genders (0.92). As before, we observed high consistency in the data, affirming the validity and efficacy of our data cleaning methodology.  

\subsubsection{Distortion score analysis}
To conduct a consistency analysis, we converted the binary distortion labels into probabilistic values by dividing the number of positive labels of each distortion by the total number of labels collected, and 
then computed the correlations between the resulting vectors. The average SRCC (inter-subject consistency) values for the distortion categories were: blurry (0.75), shaky (0.62), bright (0.68), dark (0.60), grainy (0.35), and none (0.85). Some distortion categories were harder to consistently identify than others. The high agreement on `none' shows that it is easier to determine the absence of distortions. Similarly, the SRCC values computed between image and patch distortions were: blurry (0.73), shaky (0.68), bright (0.60), dark (0.62), grainy (0.46), and none (0.73). The lower correlations for some distortions (like `dark') suggests that the perception of distortions that are globally apparent may be more weakly impacted by local quality. 

\subsubsection{MOS and distortion distributions}
\begin{figure}[t]
\begin{center}
\includegraphics[width=1\linewidth]{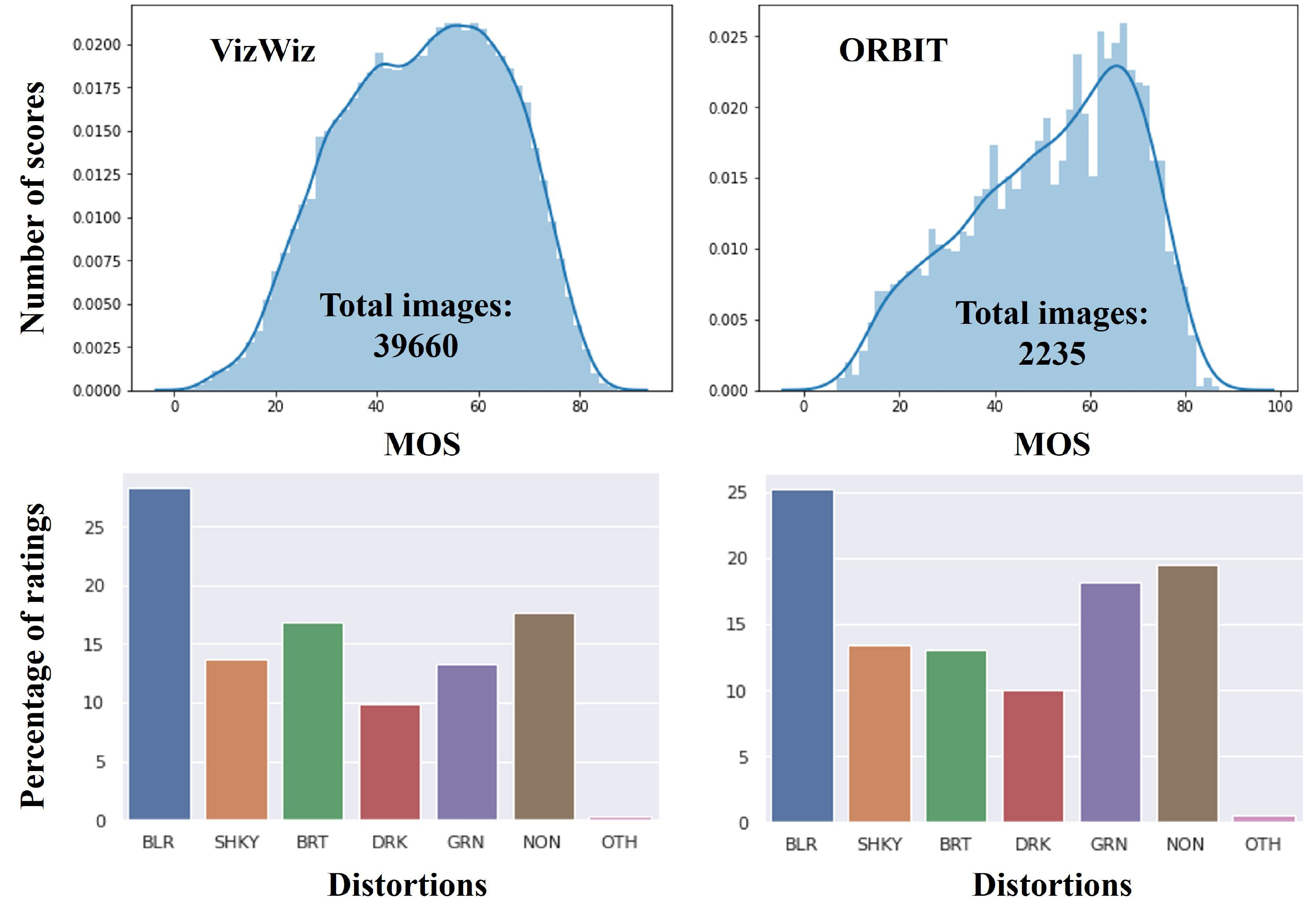}

\caption{\scriptsize{\textbf{Ground Truth MOS and distortion histograms of the two databases}. Left column is the data collected on VizWiz \cite{vizwiz} images, and right column is the data collected on ORBIT \cite{orbit} images. The plots below show the distribution of the distortions in each dataset.}}
\label{fig:MOShists}
\vspace{-1.5em}
\end{center}
\end{figure}

Fig. \ref{fig:MOShists} (top) plots the MOS distribution of the images in the VizWiz and ORBIT datasets. Fig. \ref{fig:MOShists} (bottom) shows the proportional distribution of distortion ratings in the two datasets. As expected, `blurry' was the most prominent distortion in both datasets. Overall, ORBIT contains more images of higher quality and with fewer distortions. Also, `grainy' and `dark' images, which occurred less often, are associated with less consistent ratings and are thus harder to predict. As we show in Sec.~\ref{sec:distortion_prediction}, this non-uniform distribution of distortion types makes it harder to train models that can perform equally well on all classes.    

\section{Modeling and Experiments}\label{sec:modeling}
The new database just described is intended to serve as a basic tool for developing models, and algorithms derived from them, that are able to accurately predict, and provide feedback on, the perceptual quality and distortions present in pictures. Towards realizing this aim, we have designed efficient model architectures able to predict technical quality and distortions on this kind of impaired content, while accounting for local-to-global percepts, using a  multi-task learning framework. 
\subsection{Data Pre-processing}\label{sec:preprocess}
Unlike the collected human opinion scores, which can be used in their raw form, the distortion labels had to be transformed into suitable output labels for training. Since distortion type prediction is a classification task, we considered the possibility of binarizing the labels. We considered a variety of binarization parameters. For example, the authors of \cite{vizwiz} considered $40\%$ ($2$ out of $5$) agreement to affirm the presence of a distortion. Since there were no perceptual principles to guide us, we experimented on three threshold values: $0.3$, $0.4$, and $0.5$ and assumed the top three distortions (highest proportions of the total number of labels) to be present in each analyzed image. Unfortunately, an inter-subject consistency analysis on the binarized labels found that the SRCC values dropped by $17\%$--$34\%$ as compared to those obtained on the probabilistic values in Sec. \ref{sec:data_analysis}. Similarly, we noticed a drop of $27\%$--$44\%$ of correlation between assigned patch vs image distortion classes, as compared to the probabilistic approach. Among all the binarization strategies, the `max-three' approach led to the worst correlation values. Moreover, as the thresholds were increased, the distortion distributions became more skewed, leading to worse predictions, since hard labels reduce robustness on out-of-distribution samples~\cite{robustness}. Hence, in the end, we decided to train and test all models using probabilistic labels. 

Given the two prediction tasks at hand, we studied the design of both no-reference image quality prediction models and distortion classification models, with a goal of building multi-task models capable of both tasks.
\subsection{Image Quality Models} 
\subsubsection{Architecture} The model structure we employed consists of a deep CNN backbone, followed by two-dimensional global average pooling, then two fully connected layers of size $512$ and $32$, respectively, and a single output neuron. The model was trained for $10$ epochs using the Adam optimizer with MSE loss. The learning rate was set to $5\times 10^{-4}$ for the first 5 epochs, then with a decay rate of $0.1$ per epoch. We experimented with ResNet-50V2 \cite{resnet50v2}, Xception \cite{xception}, and ResNeXt-50 \cite{resnext} backbones pre-trained on ImageNet \cite{imageNet} and fine-tuned on VizWiz images. 

\subsubsection{Dataset splits} We used the same train-validation-test split as provided by the authors of VizWiz-QualityIssues \cite{vizwiz}. The training, validation, and testing set consists of $23.9$K ($60.3\%$), $7.7$K ($19.6\%$), and $8$K ($20.1\%$) images respectively. A similar split was applied on the patch dataset.

\subsubsection{Baselines and evaluation metrics} The trained models were compared against several baselines, including shallow and deep learners (whose code was publicly available). We included the popular NR image quality prediction models, BRISQUE \cite{brisque}, NIQE \cite{niqe}, and FRIQUEE \cite{friquee}, which extract perceptually relevant statistical image features to train an SVR. We also compared against deep picture quality models such as CNNIQA \cite{cnnIqa} and NIMA \cite{nima} (with a VGG-16 \cite{vgg16} backbone and a single regressed quality score as output), the PaQ-2-PiQ baseline, and the PaQ-2-PiQ RoIPool model with backbones pre-trained on the large LIVE-FB \cite{paq2piq} dataset, then fine-tuned on our dataset. Following standard comparison methods in the field of image quality assessment, we evaluated the model performances using SRCC and LCC.

\begin{table}[t]
\captionsetup{font=scriptsize}
\setlength\extrarowheight{1.0pt}
\centering
\footnotesize
\begin{tabular}{P{3.8cm}||P{0.8cm}|P{0.8cm}||P{1.2cm}}
\hline
\textbf{Model} & \textbf{SRCC} & \textbf{LCC} & \textbf{\# params} \\
\hline
BRISQUE \cite{brisque} & 0.71 & 0.72 & - \\
NIQE \cite{niqe} & 0.68 & 0.70 & - \\
FRIQUEE \cite{friquee} & 0.72 & 0.69 & - \\
\hline
CNNIQA \cite{cnnIqa} & 0.78 & 0.79 & 0.7M  \\
NIMA \cite{nima} & 0.83 &	0.83 & 14.7M \\
P2P-Baseline (ResNet-18) \cite{paq2piq} & 0.87 & 0.88 & 11.7M  \\
P2P-RoIPool (ResNet-18) \cite{paq2piq} & \textbf{0.90} & \textbf{0.90} & 11.7M  \\
\hline
Xception & 0.86 & 0.88 & 21.9M \\
ResNeXt-50 & \textbf{0.90} & 0.89  & 24.1M\\
ResNet-50V2 & \textbf{0.90} & \textbf{0.90} & 24.6M  \\
\hline
\end{tabular}
\caption{\footnotesize{\textbf{Performance of image quality models evaluated on the new LIVE-Meta VI-UGC dataset.}} Higher values indicate better performance.}
\vspace{-1.5em}
\label{tbl:quality_models}
\end{table}

\subsubsection{Results} 
From Table~\ref{tbl:quality_models}, we note that models trained with shallow learners on extracted features yielded lower prediction accuracy than the deep models, reflecting the limited abilities of traditional features to capture complex distortions of natural images. CNNIQA \cite{cnnIqa}, which is a shallow CNN model, outperformed the traditional algorithms, but fell short of the performances of deeper models. We observed that performance generally was higher for deeper models (ResNet-50V2, ResNeXt-50, and Xception), which outperformed NIMA \cite{nima} implemented with a VGG-16 backbone. The PaQ-2-PiQ RoIPool model achieved the best performance, demonstrating the efficacy of exploiting the relationships that exist between local (patch) and global quality perception. The performances of the deeper backbones were similar (and close to human performance -- SRCC $0.93$ as observed in Sec. \ref{sec:data_analysis} ), suggesting that even heavier models would not produce better performances. 

\subsection{Distortion Prediction Models} \label{sec:distortion_prediction}

\subsubsection{Architecture and Implementation}
Because our models generate continuous probabilistic outputs, as described in Sec. \ref{sec:preprocess}, we treat distortion prediction as a regression problem. Similar to the quality model architecture, our proposed model consists of a deep CNN backbone, followed by global pooling, and two fully connected layers. Instead of producing a single output, it has seven output neurons, each expressing a score for a separate distortion class. As before, we experimented with three backbones - ResNet-50V2 \cite{resnet50v2}, Xception \cite{xception}, and ResNeXt-50 \cite{resnext}. 
The hyperparameters were kept the same, except the initial learning rate was set to $10^{-3}$. The same train-validation-test split was used. 

\subsubsection{Baselines and evaluation}
The trained models were compared against other deep models. We include two models from \cite{flickrdist}, where the authors used pre-trained Atrous VGG-16 \cite{atrvgg16} and ResNet-101 \cite{resNet} backbones with a single fully-connected head layer to predict synthetic distortions. We also tested the model \cite{vizwiz} composed of an Xception \cite{xception} backbone (pre-trained on ImageNet \cite{imageNet}), by fine-tuning the head layers only. The distortion labels and outputs lie within [$0$,$1$], and we again used SRCC to evaluate the performance.

\subsubsection{Results} As may be observed from Table \ref{tbl:dist_models}, the fine-tuned deep models outperformed the baselines. Atrous VGG-16 \cite{flickrdist} performed the worst, while ResNet-50V2 \cite{resnet50v2} and ResNeXt-50 \cite{resnext} models consistently performed best on most distortion classes. All the fine-tuned models yielded similar performances across all classes. However, the distortion distribution in the dataset is quite skewed (Fig. \ref{fig:MOShists}), hence the prediction performances varied across classes. The low performance on the `bright' and `grainy' classes is consistent with the low agreement among the subjects on these distortions (Sec. \ref{sec:data_analysis}).
\begin{table}[t]
\captionsetup{font=scriptsize}
\setlength\extrarowheight{1.0pt}
\centering
\footnotesize
\begin{tabular}{P{2.3cm}||P{0.5cm}|P{0.5cm}|P{0.5cm}|P{0.5cm}|P{0.5cm}|P{0.5cm}}

\hline
\textbf{Model} & \textbf{BLR} & \textbf{SHK} & \textbf{BRT} & \textbf{DRK} & \textbf{GRN} & \textbf{NON}\\
\hline
\hline
AtrousVGG-16 \cite{flickrdist} & 0.75 & 0.73 &  0.60 & 0.69 & 0.45 & 0.77  \\
ResNet-101 \cite{flickrdist} & 0.81 & 0.77 & \textbf{0.69} & 0.75 & 0.45 & 0.81   \\
Xception \cite{vizwiz} & 0.79 & 0.75 &  0.65 & 0.80  & \textbf{0.56} & 0.78   \\
\hline
ResNeXt-50 & \textbf{0.82} & 0.75 & 0.68 & 0.79 & \textbf{0.56} & \textbf{0.82}  \\
ResNet-50V2 & 0.81 & \textbf{0.78} &	0.67 & \textbf{0.81} & 0.50 & \textbf{0.82} \\
\hline
\end{tabular}
\caption{\footnotesize{\textbf{Performances of distortion prediction models} on the new dataset. All values are SRCC; higher values indicate better performance. 
}
}
\label{tbl:dist_models}
\vspace{-0.5em}
\end{table}


\subsection{Multi-task Models}
\subsubsection{Architecture and implementation}
\begin{figure}[htb]
\begin{center}
\includegraphics[width=1\linewidth]{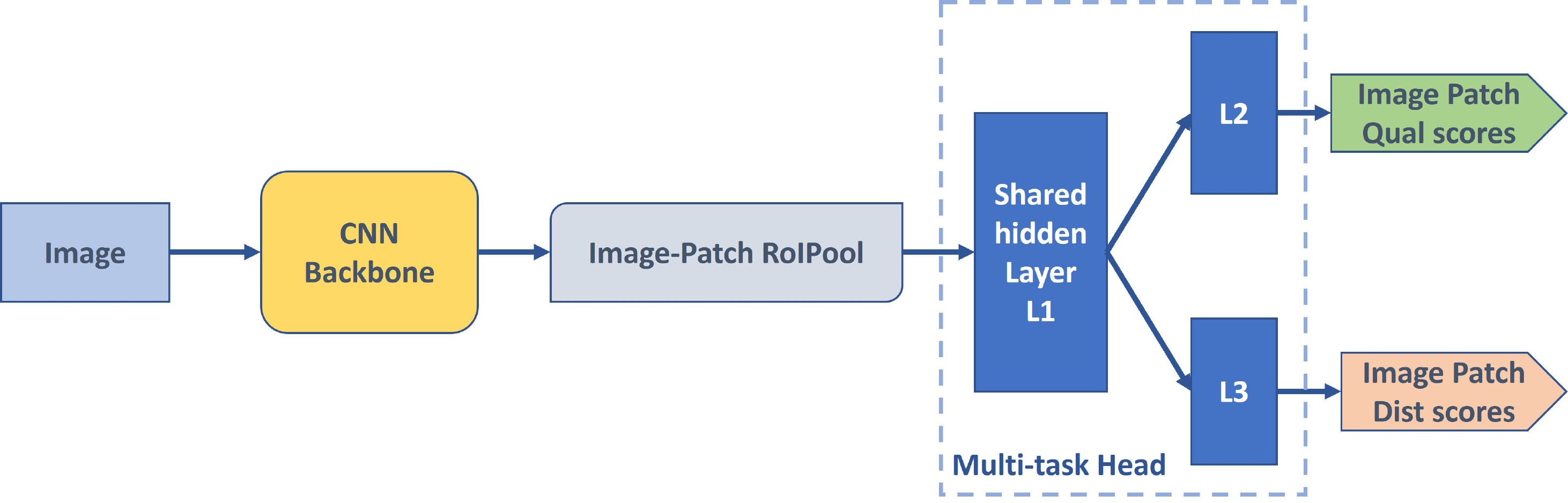}
\caption{\scriptsize{\textbf{The proposed P2P++ model} extends the PaQ-2-PiQ RoIPool \cite{paq2piq} model by including a multi-task head that simultaneously produces both quality and distortion scores, at both patch and whole image scales.}}
\vspace{-1em}
\label{fig:framework}
\end{center}
\end{figure}

Combining both tasks -- quality and distortion type predictions -- into a single model bears two advantages: a) fewer computations and faster inferencing, crucial for supplying real-time  feedback to users (Sec. \ref{sec:feedback}); and b) shared distortion and quality features can lead to better predictions \cite{qualnet}. Starting with PaQ-2-PiQ (P2P) RoIPool \cite{paq2piq} model as a base, we modified it by attaching a multi-task head to conduct both quality and distortion predictions. This multi-task model, which we call P2P++, produces quality and distortion predictions for each class, on both entire images and local patches simultaneously (Fig. \ref{fig:framework}). The head has a shared layer of size $512$, followed by two separate layers of size $32$, dedicated to separate tasks. In addition to training P2P++ (which has a ResNet-18 \cite{resNet} backbone pre-trained on LIVE-FB \cite{paq2piq}), we also experimented with ResNet-50V2 \cite{resnet50v2} and Xception \cite{xception} baselines trained on images only. The hyperparameters for training were the same as for the distortion model setup, using the same train-validation-test split. 

\subsubsection{Baselines and evaluation} We compared the performance of our models against two competitive multi-task deep models - IQACNN++ \cite{cnniqapp} and QualNet \cite{qualnet}. IQACNN++ consists of a shallow CNN backbone, while QualNet contains a VGG-16 \cite{vgg16} backbone and predicts global quality using fused distortion and quality features. 

\subsubsection{Results} From Table \ref{tbl:multitask_models}, it may be observed that the shallow IQACNN++ \cite{cnniqapp} model yielded the worst results. QualNet \cite{qualnet} was able to outperform IQACNN++, but struggled on multiple distortion categories. The larger models equipped with ResNet-50V2 and Xception backbones performed quite well, but the much lighter P2P++ model was able to achieve the best performance on almost all categories. Again, by inferencing on learned local-to-global quality and distortion features, better results were obtained at lower cost. As before, all of the models had more difficulty predicting the `bright' and `grainy' distortion types. 

\begin{table}[t]
\captionsetup{font=scriptsize}
\setlength\extrarowheight{1.0pt}
\centering
\footnotesize
\begin{tabular}{P{2cm}||P{0.43cm}|P{0.43cm}|P{0.43cm}|P{0.43cm}|P{0.43cm}|P{0.43cm}||P{0.43cm}}
\hline
\textbf{Model} & \textbf{BLR} & \textbf{SHK} & \textbf{BRT} & \textbf{DRK} & \textbf{GRN} & \textbf{NON} & \textbf{Qual}\\
\hline
\hline
IQACNN++ \cite{cnniqapp}& 0.65 & 0.52 &  0.27 & 0.57 & 0.40 & 0.62 & 0.78  \\
QualNet \cite{qualnet} & 0.70 & 0.60 & 0.46 & 0.70 & 0.29 & 0.71 & 0.81  \\
\hline
Xception & 0.78 & 0.73 &  \textbf{0.64} & 0.75  & 0.51 & \textbf{0.78} & 0.88  \\
ResNet-50V2 & 0.80 & 0.76 & 0.62 & 0.77 & 0.51 & 0.76 & \textbf{0.90} \\
P2P++ & \textbf{0.82} & \textbf{0.77} &	0.60 & \textbf{0.78} & \textbf{0.53} & \textbf{0.78} & \textbf{0.90}\\
\hline
\end{tabular}
\caption{\footnotesize{\textbf{Performance of the multi-task models} on the new subjective test dataset. All values are SRCC; higher values indicate better performance.
}
}
\label{tbl:multitask_models}
\end{table}


\subsection{Ablations}
\subsubsection{Performance on patches} 
Table \ref{tbl:patches} summarizes the quality performance of the multi-task models on patches. This is important, since giving feedback on local distortion occurences may further assist visually impaired users. P2P++ performs the best on both the random and salient patches, closely followed by ResNet-50V2. This validates the localization capabilities of the patch model. The performance on salient patches was slightly better than on random patches, perhaps because they tend to capture visibly obvious and annoying distortions that draw attention and are easier to predict. 
\begin{table}[t]
\captionsetup{font=scriptsize}
\setlength\extrarowheight{1.0pt}
\centering
\footnotesize
\begin{tabular}{P{2cm}||P{0.56cm}|P{0.56cm}||P{0.56cm}|P{0.56cm}||P{0.56cm}|P{0.56cm}}
\hline
& \multicolumn{2}{c||}{\textbf{All Patches}} & \multicolumn{2}{c||}{\textbf{Salient}} & \multicolumn{2}{c}{\textbf{Random}} \\
\hline 
\hline
\textbf{Model} & \textbf{SRCC} & \textbf{LCC} & \textbf{SRCC} & \textbf{LCC} & \textbf{SRCC} & \textbf{LCC}\\
\hline
IQACNN++ \cite{cnniqapp} & 0.71 & 0.71 &  0.71 & 0.70 & 0.72 & 0.72  \\
QualNet \cite{qualnet} & 0.77 & 0.77 & 0.78 & 0.77 & 0.77 & 0.76   \\
\hline
Xception & 0.84 & 0.84 &  0.85 & 0.84  & 0.84 & 0.83   \\
ResNet-50V2 & 0.87 & \textbf{0.87} & 0.87 & 0.87 & 0.86 & 0.86  \\
P2P++ & \textbf{0.88} & \textbf{0.87} & \textbf{0.88} & \textbf{0.88} & \textbf{0.87} & \textbf{0.87}  \\
\hline
\end{tabular}
\caption{\footnotesize{\textbf{Quality prediction results on the patches} in the new subjective dataset. Higher values indicate better performance.
}
}
\label{tbl:patches}
\vspace{-1em}
\end{table}


The distortion performances of the multi-task models on all patches are summarized in Table \ref{tbl:patch_dist}. P2P++ performed the best on average, whereas, the models with Xception \cite{xception} and ResNet-50V2 \cite{resnet50v2} backbones also performed very well. 
\begin{table}[h]
\captionsetup{font=scriptsize}
\setlength\extrarowheight{1.0pt}
\centering
\footnotesize
\begin{tabular}{P{2.8cm}||P{0.5cm}|P{0.5cm}|P{0.5cm}|P{0.5cm}|P{0.5cm}|P{0.5cm}}
\hline
\textbf{Model} & \textbf{BLR} & \textbf{SHK} & \textbf{BRT} & \textbf{DRK} & \textbf{GRN} & \textbf{NON} \\
\hline
\hline
IQACNN++ \cite{cnniqapp} & 0.52  & 0.31 & 0.28 & 0.67 & 0.32 & 0.58\\
QualNet \cite{qualnet} & 0.68 & 0.49 & 0.39 & 0.71 & 0.47 & 0.59\\
\hline
Xception & 0.76 & 0.62 &  \textbf{0.48} & 0.73  & \textbf{0.51} & 0.74\\
ResNet-50V2 & 0.80 & 0.62 & 0.46 & \textbf{0.75} & 0.48 & 0.74 \\
P2P++ (ResNet-18) & \textbf{0.81} & \textbf{0.65} & 0.42 & \textbf{0.75} & 0.50 & \textbf{0.76}\\
\hline
\end{tabular}
\caption{\footnotesize{\textbf{Distortion prediction performance of the multi-task models on all patches} in the new subjective test dataset. All values are SRCC; higher values indicate better performance.
}
}
\label{tbl:patch_dist}
\end{table}



\subsubsection{Prediction on different MOS ranges}
To analyze the performances of the models on images, we divided the MOS range into three equal non-overlapping ranges: $0-33$ ($18.4\%$ of images), $34-66$ ($64.6\%$ of images), and $67-100$ ($17\%$ of images), representing low, medium, and high quality images, respectively. Table \ref{tbl:mos_range_quality} shows the quality prediction performances of the different multi-task models on separate MOS ranges. It was observed that most models performed well in the middle range (medium quality images) compared to the low and high MOS ranges. This is likely due to the higher number of medium quality images in our dataset compared to the two extremes.

\begin{table*}[htb]
\captionsetup{font=scriptsize}
\setlength\extrarowheight{1.0pt}
\centering
\footnotesize
\begin{tabular}{P{2.5cm}||P{0.6cm}|P{0.6cm}||P{0.6cm}|P{0.6cm}||P{0.6cm}|P{0.6cm}||P{0.6cm}|P{0.6cm}}
\hline
& \multicolumn{2}{c||}{\textbf{Low MOS}} & \multicolumn{2}{c||}{\textbf{Medium MOS}} & \multicolumn{2}{c||}{\textbf{High MOS}} & \multicolumn{2}{c}{\textbf{All MOS}}\\
\hline 
\hline
\textbf{Model} & \textbf{SRCC} & \textbf{LCC} & \textbf{SRCC} & \textbf{LCC} & \textbf{SRCC} & \textbf{LCC} & \textbf{SRCC} & \textbf{LCC}\\
\hline
IQACNN++ \cite{cnniqapp} & 0.54 & 0.55 &  0.60 & 0.60 & 0.21 & 0.22 & 0.78 & 0.79 \\
QualNet \cite{qualnet} & 0.57 & 0.52 & 0.62 & 0.62 & 0.42 & 0.66 & 0.81 & 0.82 \\
\hline
Xception & 0.69 & 0.67 &  0.70 & 0.72  & 0.53 & 0.80 & 0.88 & 0.87   \\
ResNet-50V2 & \textbf{0.72} & \textbf{0.73} & 0.74 & 0.73 & \textbf{0.55} & 0.81 & \textbf{0.90} & \textbf{0.90}  \\
P2P++ (ResNet-18) & 0.70 & 0.72 & \textbf{0.75} & \textbf{0.74} & \textbf{0.55} & \textbf{0.83} & \textbf{0.90} & \textbf{0.90}  \\
\hline
\end{tabular}
\caption{\footnotesize{\textbf{Quality prediction results on different disjoint MOS ranges} of the new LIVE-Meta VI-UGC dataset. Higher values indicate better performance.
}
}
\label{tbl:mos_range_quality}
\vspace{-1em}
\end{table*}


\subsubsection{Mobile-friendly version}
In a guided feedback system, fast inference is necessary. Hence, we also implemented an efficient lighter version of P2P++, where, instead of a ResNet-18 \cite{resNet}, we used a MobileNetV2 \cite{mobileNetV2} backbone and reduced the RoIPool output size to $1 \times 1$. We noticed a drop of $8.9\%$ in quality prediction and $8.6\%$ in distortion prediction performances when compared to P2P++ (with ResNet-18 backbone). However, the MobileNetV2 version contains only $1/4$ as many parameters compared to P2P++, making it much more desirable for mobile implementations.

\subsubsection{Failure Cases}
\begin{figure}[t]
\begin{center}
\includegraphics[width=1\linewidth]{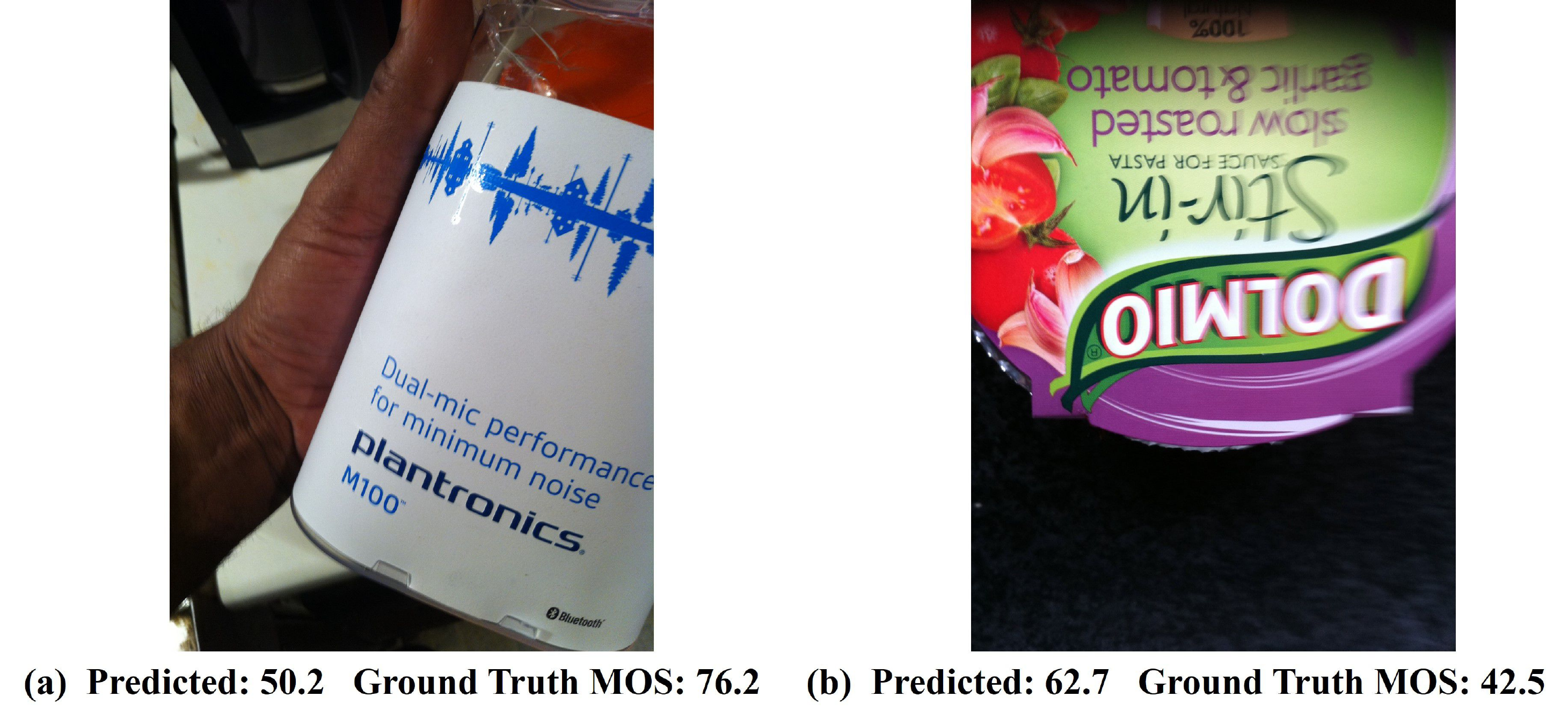}
\caption{\footnotesize{\textbf{Failure cases:} VizWiz images where predictions differed the most from human quality scores.}}
\vspace{-0.3in}
\label{fig:failure_eg}
\end{center}
\end{figure}

Fig. \ref{fig:failure_eg} (a) was rated high (MOS = $76.2$) by the humans, but obtained a low predicted score (MOS = $50.2$) from P2P++. Perhaps the blurry ``bokeh" effect on regions around the hand and background was less noticeable to the human raters than the high quality (salient) foreground. The image Fig. \ref{fig:failure_eg} (b) was rated as worse (MOS = $42.5$) by the subjects than by P2P++ (MOS = $62.7$). The non-uniformity of the blur across the image also could have caused this discrepancy. While the sharp and distorted regions of the image are of roughly equal areas, the distortions were likely more salient to the human subjects, causing them to rate it severely. The same regions were likely predicted as less salient by P2P++. These results suggest that more work needs to be done on understanding the interplay between saliency, distortion annoyance, and bokeh. 

\section{Applications of the proposed model} \label{sec:modelExt}
The models described in Sec. \ref{sec:modeling} can be extended to provide visualization and feedback to directly assist visually impaired users, as we describe next.
\subsection{Predicting Quality and Distortion Maps} \label{sec:qualityMaps}
The P2P++ model can be used to compute both spatial quality maps and distortion classification maps. Since it is trained on both global and local patch labels, it is flexible enough to compute quality predictions and distortion type predictions on any number and sizes of image patches. Inspired by \cite{paq2piq}, we utilized these outputs to create perceptual quality and distortion classification maps that span the entire image space. To generate spatial quality maps, each image is divided into non-overlapping patches of size $N\times N$, on which predicted quality scores are obtained from the model output on every patch. Similarly, on each patch, a predicted distortion vector is obtained, with multiple values corresponding to each distortion type. Each distortion output can be used to generate a corresponding distortion-specific map. The patch size ($N$) is easily varied, allowing the generation of finer or coarser maps.

\begin{figure}[htb]
\vspace{-1em}
\begin{center}
    \includegraphics[width=1\linewidth]{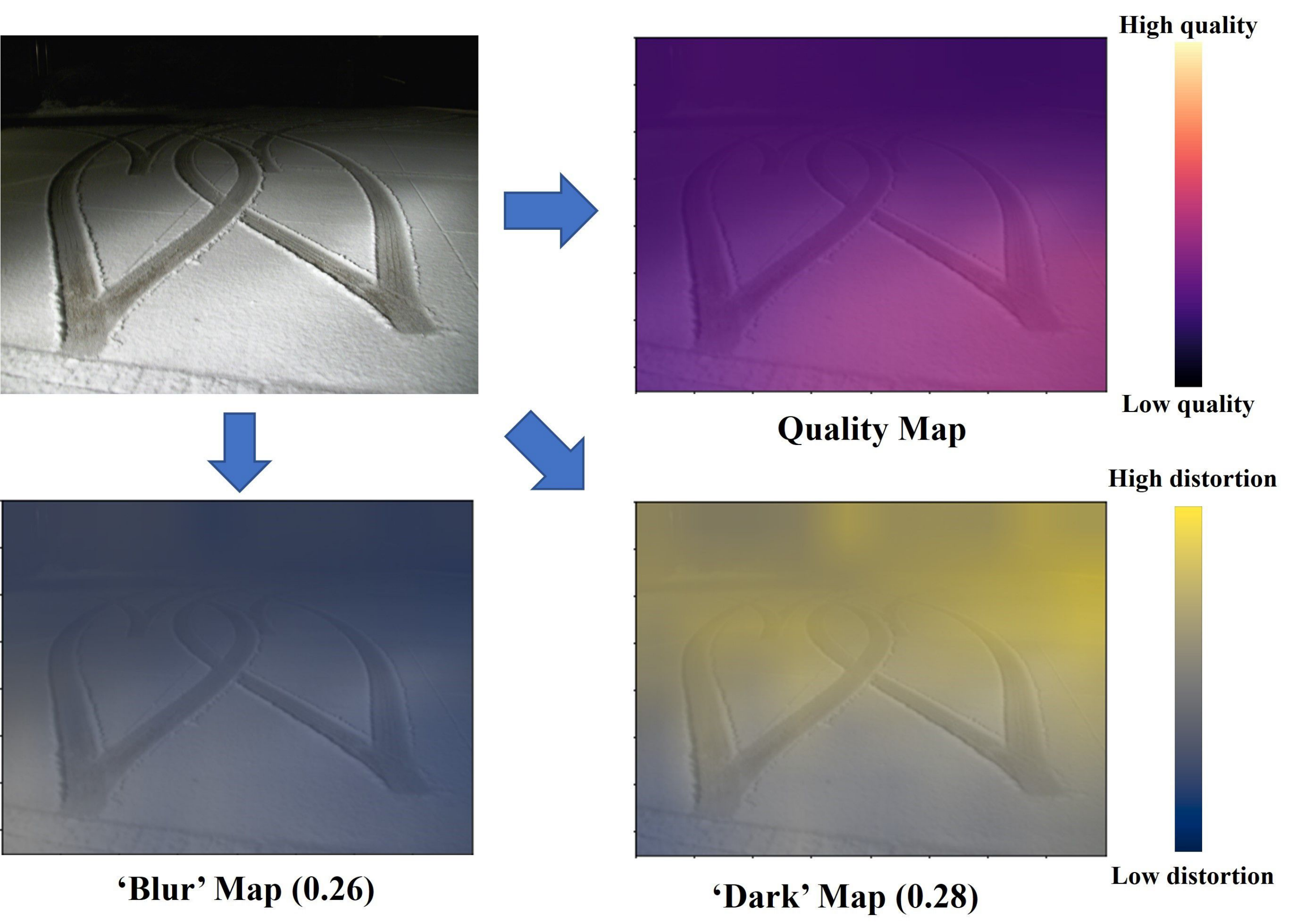}
    \captionof{figure}{\footnotesize{\textbf{Spatial quality and distortion maps:} Predicted perceptual quality and distortion maps were generated on a sample image from the new LIVE-Meta VI-UGC database \ref{sec:qualityMaps}. The top-right image shows the predicted perceptual quality map (blended with the original image using a magma colormap). The bottom two images show the `blur' and `dark' distortion maps (and their global scores) blended with a cividis colormap. Best viewed in color.}}
    \label{fig:space_quality_dist}
\end{center}
\vspace{-2.5em}
\end{figure}

Fig \ref{fig:space_quality_dist} shows the predicted quality and distortion maps (for the two most prominent predicted distortions) computed on a sample test image. The quality map accurately predicted the bottom-right part of the image to be of the highest quality, while the distortion maps predicted the bottom-left area to be blurry, and the topmost region of the image to be underexposed. This example shows the interplay between perceived quality and distortion localization.  

\subsection{Feedback to Assist Visually Impaired Users} \label{sec:feedback}
\subsubsection{Guided Photography}
\begin{figure}[htb]
\vspace{-1em}
\begin{center}
    \includegraphics[width=1\linewidth]{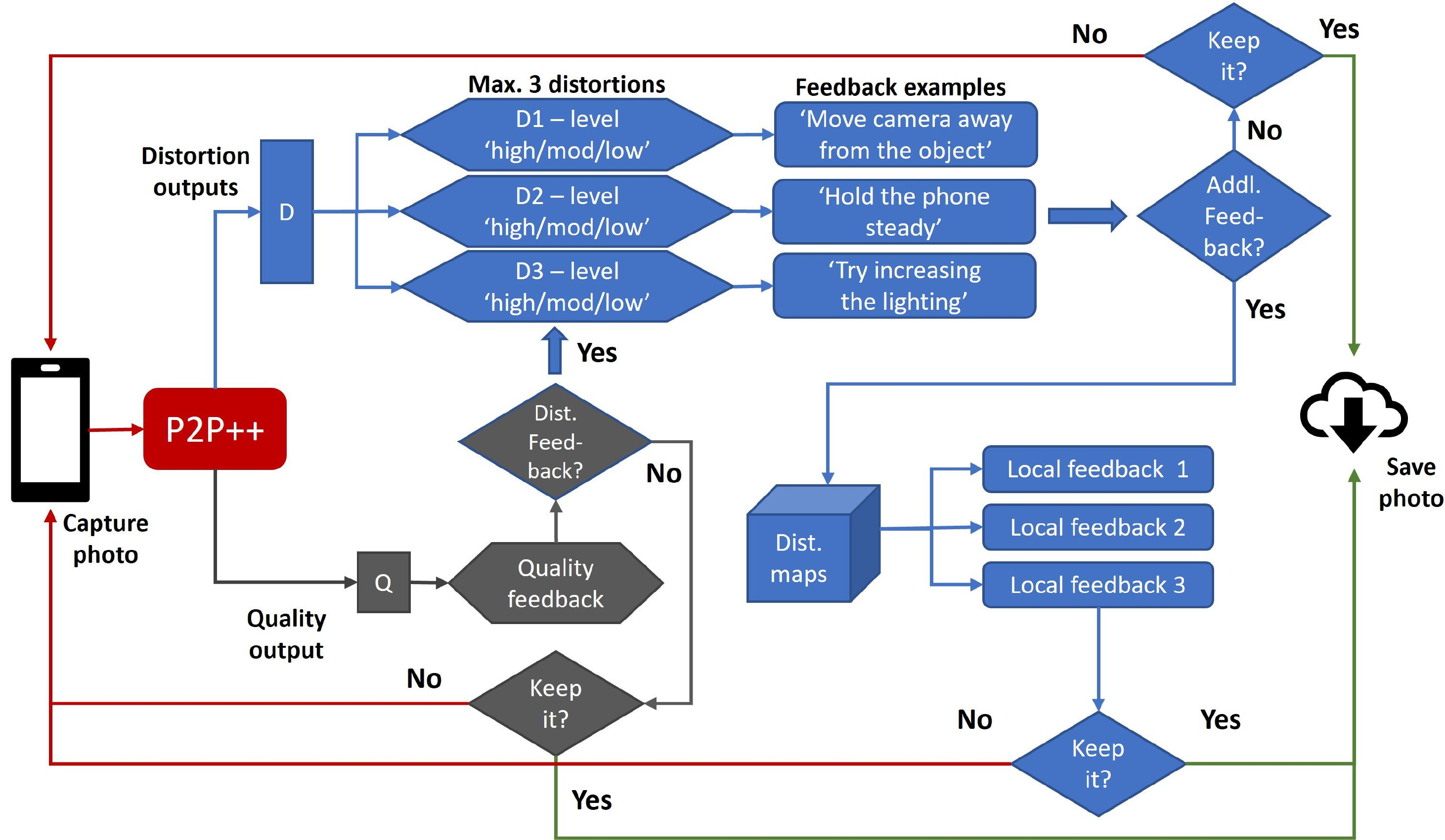}
    \vspace{-0.5em}
    \captionof{figure}{\footnotesize{\textbf{Guided Photography Framework:} Flowchart of the assistive photography framework (Sec. \ref{sec:feedback}), showing the series of prompts and advice given to guide visually impaired users, from capturing through saving a satisfactory photo. Here, `mod' stands for `moderate distortion'.
    }}
    \label{fig:guide_feedback}
\end{center}
\vspace{-1em}
\end{figure}

There are many ways of providing feedback to visually impaired users to try to help the take pictures of better perceived technical quality. This is a very challenging, multi-dimensional, and human-oriented problem, which will, over time, require strong engagement with the community of visually impaired people, and extensive ergonomic and validation studies involving visually impaired volunteer subjects. As a demonstration of the potential of our concept, we built a prototype early-stage, guided feedback system to show how our work can be used to assist visually impaired users to take better photos. The current implementation of our framework is illustrated in Fig. \ref{fig:guide_feedback}. The assistive model has two parts, a quality feedback loop, and a distortion feedback loop. The high-level, immediate model outputs are approximate English expressions of the global picture quality prediction, and of the predicted distortion levels. Specifically, the user is provided an image rating from among `Bad' (0-20), `Poor' (20-40), `Fair' (40-60), `Good' (60-80), and `Excellent' (80-100). If the user is satisfied with the quality, he/she can choose to save it, or otherwise ask for distortion feedback. In our current prototype, which is implemented on a workstation (but see below for parallel work), the feedback is given by output text; naturally, transcribed audio expressions would be used in practice. If the quality is substandard, then further feedback is required to make the application useful. If feedback on the distortion is requested, the user is informed of the three major distortions determined to be present in the image, along with the severity of each: High ($> 0.50$), Moderate ($0.20 - 0.50$), and Low ($< 0.20$). 

Based on the nature and severities of the distortions detected, our system also suggests simple ways (\textbf{base feedback}) to mitigate them. 
The following feedbacks (suggested actions) are given to the visually impaired users to mitigate those distortions predicted to be present:
\begin{itemize}
\item \textbf{`Blur':} `The phone may be too close to the object, move it away from it.'
\item \textbf{`Shaky':} `Hold the phone and the object steady.'
\item \textbf{`Bright':} `Scene is too bright' + `Try turning off the flash' / `Find proper lighting if you can.'
\item \textbf{`Dark':} `Scene is too dark, try turning on the flash or switch on the lights.'
\item \textbf{`Noisy':} `Try increasing the lighting or move the camera further from the subject.'
\item \textbf{`None':} `No major distortions seem to be present.'
\end{itemize}

As the user becomes more adept at the P2P++ system, they will be able to request and take advantage of additional, more \textbf{detailed (localized) feedback} on the picture distortions. To facilitate this, P2P++ also generates $3\times 3$ distortion maps for the three most dominant impairments, and informs the user of their relative locations in the image (top-left, bottom-right, center, etc.), as also depicted in Fig. \ref{fig:guide_feedback}.

\subsubsection{Automated Photography}
\begin{figure}[htb]
\vspace{-1em}
\begin{center}
    \includegraphics[width=0.99\linewidth]{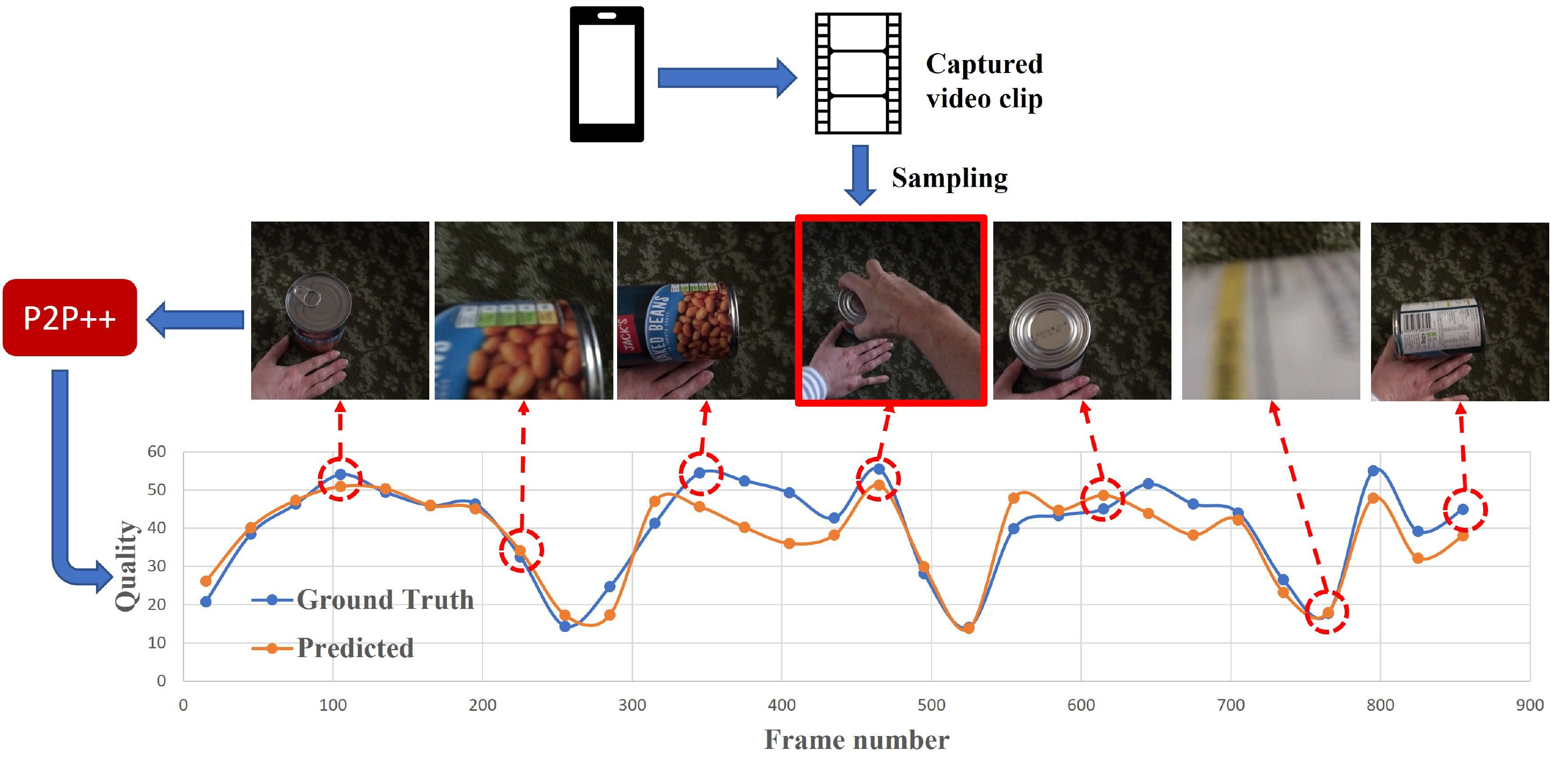}
    \vspace{-0.5em}
    \captionof{figure}{\footnotesize{\textbf{Automated photography:} The outputs of P2P++ on a sequence are used to determine the highest quality frame among a temporally sampled ORBIT~\cite{orbit} video (highlighted in red). Best viewed in color.}}
    \label{fig:auto_feedback}
\end{center}
\vspace{-1em}
\end{figure}

Although guided photography promises to be a transformative technology, we acknowledge that much work remains on developing successful feedback languages and interfaces, which in term will require working with visually impaired subjects to test and advance future systems. In the interim, there are more immediate ways to assist visually impaired users to take better pictures via simpler, albeit less comprehensive applications, which can automatically help them take better quality pictures. This can be accomplished by capturing a short video clip of the scene the subject is trying to photograph, that includes and is approximately centered at the moment the `shutter button' is depressed. By using a broad sampling of a single frame per second, a fairly wide range of qualities may be presented. Given the sampled frames, P2P++ then computes the global quality of each to determine the frame having the highest perceptual quality. The user is provided feedback on this quality (`Poor' to `Excellent') and given the option to save or discard the image. A simple demonstration on an ORBIT \cite{orbit} video is shown in Fig. \ref{fig:auto_feedback}. 

\subsubsection{Performance on ORBIT} \label{sec:cross_data}
To study the generalizability of our model, and the representation capabilities of our dataset, we also sought to test P2P++ on other, independent VI-UGC datasets. Since we could not find any such datasets, we evaluated and compared the multi-task models on a special-purpose excerpt we created from the ORBIT dataset, consisting of frames sampled from ORBIT \cite{orbit} videos. As may be observed from Table \ref{tbl:orbit_results}, P2P++ performs very well, and generally better than the much heavier ResNet-50V2 and Xception models. All of our models outperformed other multi-task models when trained on our new database and tested on the excerpted ORBIT dataset. Attaining such high performance on most distortion classes on ORBIT validates the generalizability of P2P++ to other VI-UGC media. The lower performance (of all models) on the `bright' and `grainy' categories is again due to subject ambiguity on these classes. Fig. \ref{fig:auto_feedback} illustrates the actual performance and outputs produced by P2P++ when compared to the ground truth quality scores obtained on an ORBIT video.


\begin{table}[t]
\captionsetup{font=scriptsize}
\setlength\extrarowheight{1.0pt}
\centering
\footnotesize
\begin{tabular}{P{2cm}||P{0.43cm}|P{0.43cm}|P{0.43cm}|P{0.43cm}|P{0.43cm}|P{0.43cm}||P{0.41cm}}
\hline
\textbf{Model} & \textbf{BLR} & \textbf{SHK} & \textbf{BRT} & \textbf{DRK} & \textbf{GRN} & \textbf{NON} & \textbf{Qual}\\
\hline
\hline
IQACNN++ \cite{cnniqapp}& 0.56 & 0.37 &  0.06 & 0.83 & 0.05 & 0.38 & 0.78  \\
QualNet \cite{qualnet} & 0.69 & 0.59 & 0.11 & 0.79 & 0.17 & 0.70 & 0.85  \\
\hline
Xception & 0.70 & 0.64 &  0.27 & 0.81  & \textbf{0.37} & \textbf{0.72} & 0.83  \\
ResNet-50V2 & 0.68 & 0.69 & 0.17 & \textbf{0.84} & 0.18 & 0.65 & \textbf{0.86} \\
P2P++ & \textbf{0.72} & \textbf{0.71} &	\textbf{0.30} & 0.83 & \textbf{0.37} & \textbf{0.72} & \textbf{0.86}\\
\hline
\end{tabular}
\caption{\footnotesize{\textbf{Performance of the multi-task models} when trained on our new dataset and tested on the excerpted ORBIT dataset \textbf{without fine-tuning}. All values are SRCC; higher values indicate better performance. 
}
}
\label{tbl:orbit_results}
\vspace{-1em}
\end{table}


\subsubsection{Smartphone Application}
In parallel with this work, we supervised a 5-member undergraduate Honors Senior Design team at UT-Austin over two semesters, who implemented the P2P++ MobileNetV2 version into a functioning application, available on both iOS and Android, using tools available in the SDK. The app includes an easy to use user interface and more detailed feedback expressions that is communicated via the phone's audio speaker. A demo of the application has been made available online at \cite{appDemo}. 

\section{Concluding Remarks}\label{sec:conclusion}
The success of computer vision algorithms can be largely measured by the benefits granted to ordinary people to enhance their quality of life. To that end, assisting visually impaired people to take better quality pictures can give them more prominent voices on social media platforms, and can also assist them with other visual tasks such as recognition and captioning. Assessing perceptual quality and distortions on VI-UGC is a difficult, but important and little-addressed problem. Our work makes substantive progress towards that goal by the proposed VI-UGC targeted dataset, a VI-UGC quality and distortion prediction model, and a prototype system that supplies specialized feedback to help guide, assist, automate, and improve their photographic efforts. Of course, while we believe that this work is a step in the right direction, this field is still nascent with very significant challenges remaining. 


\section*{Acknowledgments}
This research was sponsored by a grant from Meta AI Research, and by grant number 2019844 from the National Science Foundation AI Institute for Foundations of Machine Learning (IFML). The authors thank The University of Texas at Austin for providing computational resources that have contributed to the research results reported in this paper.



 

{\small
\bibliographystyle{IEEEtran}
\bibliography{egbib}
}

\vfill

\end{document}